\pdfoutput=1

\documentclass[10pt,journal,compsoc]{IEEEtran}
\usepackage{lipsum}
\usepackage{soul}
\usepackage{caption}
\usepackage{subcaption}
\usepackage{graphicx}

\usepackage{color}
\usepackage{multirow}
\usepackage{booktabs} 
\usepackage{colortbl}

\ifCLASSOPTIONcompsoc
\usepackage[nocompress]{cite}
\else
\usepackage{cite}
\fi
\ifCLASSINFOpdf
\else
\fi
\usepackage{amsmath}
\usepackage{array}

\usepackage[strings]{underscore}
\usepackage{ragged2e}

\begin{document}
	%
	\title{Context Based Emotion Recognition using EMOTIC Dataset}
	%
	%
	%
	%

	\author{Ronak~Kosti,
		Jose~M.~Alvarez,
		Adria~Recasens,
		Agata~Lapedriza%
		\IEEEcompsocitemizethanks{\IEEEcompsocthanksitem Ronak Kosti \& Agata Lapedriza are with Universitat Oberta de Catalunya, Spain. {Email: rkosti@uoc.edu, alapedriza@uoc.edu}. 
			\IEEEcompsocthanksitem Adria Recasens is with the Computer Science and Artificial Intelligence Laboratory, Massachusetts Institute of Technology, USA. {Email: recasens@mit.edu}. 
			\IEEEcompsocthanksitem Jose M. Alvarez is with NVIDIA, USA. {Email: jalvarez.research@gmail.com}
			\IEEEcompsocthanksitem Project Page: http://sunai.uoc.edu/emotic/ }}
	
	\markboth{IEEE TRANSACTIONS ON PATTERN ANALYSIS AND MACHINE INTELLIGENCE}{Kosti \MakeLowercase{\text it{et al.}}: Emotion Recognition in Scene Context using EMOTIC Dataset}%
	
	\IEEEtitleabstractindextext{%
		\begin{abstract}
			\justifying	
			In our everyday lives and social interactions we often try to perceive the emotional states of people. There has been a lot of research in providing machines with a similar capacity of recognizing emotions. From a computer vision perspective, most of the previous efforts have been focusing in analyzing the facial expressions and, in some cases, also the body pose. Some of these methods work remarkably well in specific settings. However, their performance is limited in natural, unconstrained environments. Psychological studies show that the scene context, in addition to facial expression and body pose, provides important information to our perception of people's emotions. However, the processing of the context for automatic emotion recognition has not been explored in depth, partly due to the lack of proper data. In this paper we present EMOTIC, a dataset of images of people in a diverse set of natural situations, annotated with their apparent emotion. The EMOTIC dataset combines two different types of emotion representation: (1) a set of 26 discrete categories, and (2) the continuous dimensions \emph{Valence}, \emph{Arousal}, and \emph{Dominance}. We also present a detailed statistical and algorithmic analysis of the dataset along with annotators' agreement analysis. Using the EMOTIC dataset we train different CNN models for emotion recognition, combining the information of the bounding box containing the person with the contextual information extracted from the scene. Our results show how scene context provides important information to automatically recognize emotional states and motivate further research in this direction. Dataset and Code is open-sourced and available on \textit{https://github.com/rkosti/emotic}. 
			Link for the published article \textit{https://ieeexplore.ieee.org/document/8713881}.
		\end{abstract}
		
		\begin{IEEEkeywords}
			Emotion recognition, Affective computing, Pattern recognition
	\end{IEEEkeywords}}

	\maketitle

	\IEEEdisplaynontitleabstractindextext

	%
	\IEEEpeerreviewmaketitle
	\IEEEraisesectionheading{\section{Introduction}\label{intro}}
	
	\IEEEPARstart{O}ver the past years, the interest in developing automatic systems for recognizing  emotional states has grown rapidly. We can find several recent works showing how emotions can be inferred from cues like text \cite{contextText}, voice \cite{contextVoice}, or visual information \cite{contextBody,contextFace}. The automatic recognition of emotions has a lot of applications in environments where machines need to interact or monitor humans. For instance, automatic tutors in an online learning platform would provide better feedback to a student according to her level of motivation or frustration. Also, a car with the capacity of assisting a driver can intervene or give an alarm if it detects the driver is tired or nervous. 

In this paper we focus on the problem of emotion recognition from visual information. Concretely, we want to recognize the apparent emotional state of a person in a given image. This problem has been broadly studied in computer vision mainly from two perspectives: (1) facial expression analysis, and (2) body posture and gesture analysis. Section \ref{related_work} gives an overview of related work on these perspectives and also on some of the common public datasets for emotion recognition. 

\begin{figure}
	\centering
	\includegraphics[width=2.5in]{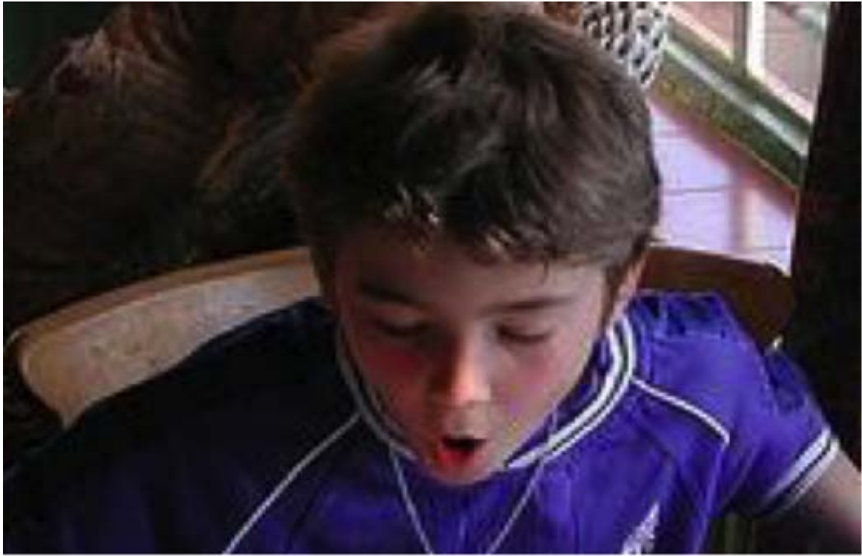}
	\caption{How is this kid feeling? Try to recognize his emotional states from the person bounding box, without scene context.}
	\label{fig_teaser}
\end{figure}

\begin{figure}
	\centering
	\includegraphics[width=\linewidth]{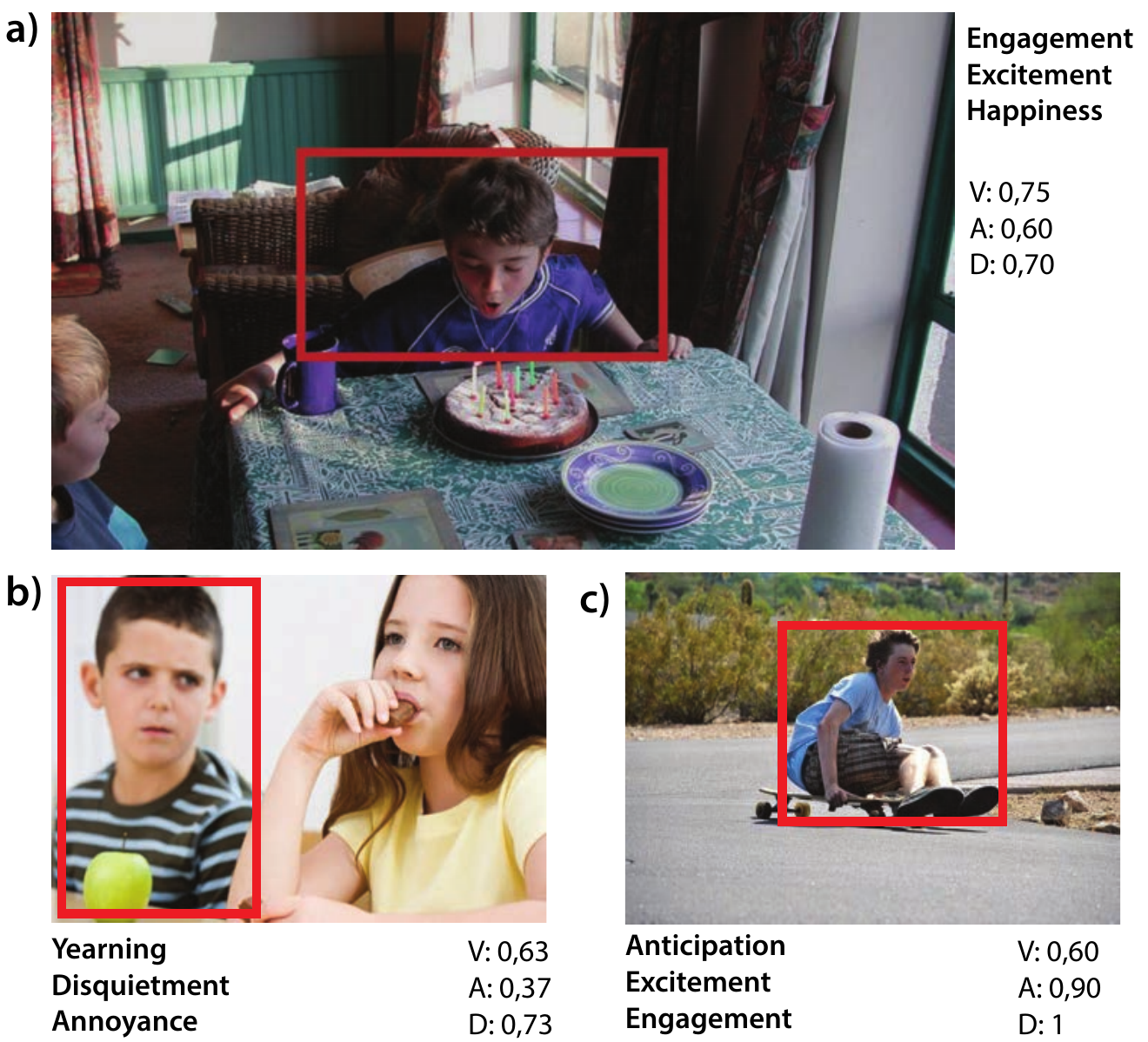}
	\caption{Sample images in the EMOTIC dataset along with their annotations.}
	\label{example_annotations}
\end{figure}

Although face and body pose give lot of information on the affective state of a person, our claim in this work is that scene context information is also a key component for understanding emotional states. Scene context includes the surroundings of the person, like the place category, the place attributes, the objects, or the actions occurring around the person. Fig. \ref{fig_teaser} illustrates the importance of scene context for emotion recognition. When we just see the kid it is difficult to recognize his emotion (from his facial expression it seems he is feeling $Surprise$). However, when we see the context (Fig.~\ref{example_annotations}.a) we see the kid is celebrating his birthday, blowing the candles, probably with his family or friends at home. With this additional information we can interpret much better his face and posture and recognize that he probably feels \emph{engaged}, \emph{happy} and \emph{excited}.

The importance of context in emotion perception is well supported by different studies in psychology \cite{barrett2011context,barrett2017emotions}. In general situations, facial expression is not sufficient to determine the emotional state of a person, since the perception of the emotion is heavily influenced by different types of context, including the scene context \cite{contextVoice,contextBody,contextFace}.

In this work, we present two main contributions. Our first contribution is the creation and publication of the \textbf{EMOTIC} (from EMOTions In Context) Dataset. The EMOTIC database is a collection of images of people annotated according to their apparent emotional states. Images are spontaneous and unconstrained, showing people doing different things in different environments. Fig. \ref{example_annotations} shows some examples of images in the EMOTIC database along with their corresponding annotations. As shown, annotations combine $2$ different types of emotion representation: Discrete Emotion Categories and $3$ Continuous Emotion Dimensions \emph{Valence}, \emph{Arousal}, and \emph{Dominance} \cite{pad_model}. The EMOTIC dataset is now publicly available for download at the EMOTIC website\footnote{http://sunai.uoc.edu/emotic/\label{web_emotic}}. Details of the dataset construction process and dataset statistics can be found in section \ref{emotic_dataset}.

Our second contribution is the creation of a baseline system for the task of emotion recognition in context. In particular, we present and test a Convolutional Neural Network (CNN) model that jointly processes the window of the person and the whole image to predict the apparent emotional state of the person. Section \ref{emotic_recognition} describes the CNN model and the implementation details while section \ref{experiments} presents our experiments and discussion on the results. All the trained models resulting from this work are also publicly available at the EMOTIC website$^{\ref{web_emotic}}$.

This paper is an extension of the conference paper "Emotion Recognition in Context", presented at the IEEE International Conference on Computer Vision and Pattern Recognition (CVPR) 2017 \cite{emotic}. We present here an extended version of the EMOTIC dataset, with further statistical dataset analysis, an analysis of scene-centric algorithms on the data, and a study on the annotation consistency among different annotators. This new release of the EMOTIC database contains $44.4\%$ more annotated people as compared to its previous smaller version. With the new extended dataset we retrained all the proposed baseline CNN models with additional loss functions. We also present comparative analysis of two different scene context features, showing how the context is contributing to recognize emotions in the wild.  

\section{Related Work}
\label{related_work}
Emotion recognition has been broadly studied by the Computer Vision community. Most of the existing work has focused on the analysis of facial expression to predict emotions \cite{pantic2000expert,li2009facial}. The base of these methods is the \emph{Facial Action Coding System} \cite{friesen1978facial}, which encodes the facial expression using a set of specific localized movements of the face, called \emph{Action Units}. These facial-based approaches \cite{pantic2000expert,li2009facial} usually use facial-geometry based features or appearance features to describe the face. Afterwards, the extracted features are used to recognize \emph{Action Units} and the basic emotions proposed by Ekman and Friesen \cite{ekman1971constants}: \emph{anger}, \emph{disgust}, \emph{fear}, \emph{happiness}, \emph{sadness}, and \emph{surprise}. Currently, state-of-the-art systems for emotion recognition from facial expression analysis use CNNs to recognize emotions or Action Units \cite{benitez2016emotionet}. 

In terms of emotion representation, some recent works based on facial expression \cite{soleymani2016continuousface} use the continuous dimensions of the \emph{$VAD$ Emotional State Model} \cite{pad_model}. The VAD model describes emotions using 3 numerical dimensions: \emph{\textbf{Valence}} (V), that measures how positive or pleasant an emotion is, ranging from \emph{negative} to \emph{positive}; \emph{\textbf{Arousal}} (A), that measures the agitation level of the person, ranging from \emph{non-active / in calm} to \emph{agitated / ready to act}; and \emph{\textbf{Dominance}} (D) that measures the level of control a person feels of the situation, ranging from \emph{submissive / non-control} to \emph{dominant / in-control}. On the other hand, Du et al. \cite{du2014compound} proposed a set of $21$ facial emotion categories, defined as different combinations of the basic emotions, like `happily surprised' or `happily disgusted'. With this categorization the authors can give a fine-grained detail about the expressed emotion. 


Although the research in emotion recognition from a computer vision perspective is mainly focused in the analysis of the face, there are some works that also consider other additional visual cues or multimodal approaches. For instance, in \cite{nicolaou2011continuous} the location of shoulders is used as additional information to the face features to recognize basic emotions. More generally, Schindler et al. \cite{schindler2008recognizing} used the body pose to recognize $6$ basic emotions, performing experiments on a small dataset of non-spontaneous poses acquired under controlled conditions. Mou et al. \cite{mou2015group} presented a system of affect analysis in still images of groups of people, recognizing group-level arousal and valence from combining face, body and contextual information. 


Emotion Recognition in Scene Context and Image Sentiment Analysis are different problems that share some characteristics. Emotion Recognition aims to identify the emotions of a person depicted in an image. Image Sentiment Analysis consists of predicting what a person will feel when observing a picture. This picture does not necessarily contain a person. When an image contains a person, there can be a difference between the emotions experienced by the person in the image and the emotions felt by observers of the image. For example, in the image of Figure \ref{example_annotations}.b, we see a kid who seems to be annoyed for having an apple instead of chocolate and another who seems happy to have chocolate. However, as observers, we might not have any of those sentiments when looking at the photo. Instead, we might think the situation is not fair and feel disapproval. Also, if we see an image of an athlete that has lost a match, we can recognize the athlete feels sad. However, an observer of the image may feel happy if the observer is a fan of the team that won the match.



\subsection{Emotion Recognition Datasets}
Most of the existing datasets for emotion recognition using computer vision are centered in facial expression analysis. For example, the GENKI database \cite{genki} contains frontal face images of a single person with different illumination, geographic, personal and ethnic settings. Images in this dataset are labelled as \textit{smiling} or \textit{non-smiling}. Another common facial expression analysis dataset is the ICML Face-Expression Recognition dataset \cite{icml}, that contains $28,000$ images annotated with $6$ basic emotions and a neutral category. On the other hand, the UCDSEE dataset \cite{ucdsee} has a set of $9$ emotion expressions acted by $4$ persons. The lab setting is strictly kept the same in order to focus mainly on the facial expression of the person.

The dynamic body movement is also an essential source for estimating emotion. Studies such as \cite{uclic1,uclic2} establish the relationship between affect and body posture using as ground truth the base-rate of human observers. The data consist of a spontaneous set of images acquired under a restrictive setting (people playing Wii games). The GEMEP database \cite{gemep} is multi-modal (audio and video) and has $10$ actors playing $18$ affective states. The dataset has videos of actors showing emotions through acting. Body pose and facial expression are combined. 

The Looking at People (LAP) challenges and competitions \cite{chalearn} involve specialized datasets containing images, sequences of images and multi-modal data. The main focus of these datasets is the complexity and variability of human body configuration which include data related to personality traits (spontaneous), gesture recognition (acted), apparent age recognition (spontaneous), cultural event recognition (spontaneous), action/interaction recognition and human pose recognition (spontaneous). 

The Emotion Recognition in the Wild (EmotiW) challenges \cite{emotiw} host 3 databases: (\emph{1}) The \textit{AFEW} database \cite{afew} focuses on emotion recognition from video frames taken from movies and TV shows, where the actions are annotated with attributes like name, age of actor, age of character, pose, gender, expression of person, the overall clip expression and the basic $6$ emotions and a neutral category; (\emph{2}) The \textit{SFEW}, which is a subset of AFEW database containing images of face-frames annotated specifically with the $6$ basic emotions and a neutral category; and (\emph{3}) the \textit{HAPPEI} database \cite{happei}, which addresses the problem of group level emotion estimation. Thus, \cite{happei} offers a first attempt to use context for the problem of predicting happiness in groups of people. 

Finally, the COCO dataset has been recently annotated with object attributes \cite{patterson2016coco}, including some emotion categories for people, such as \emph{happy} and \emph{curious}. These attributes show some overlap with the categories that we define in this paper. However, COCO attributes are not intended to be exhaustive for emotion recognition, and not all the people in the dataset are annotated with affect attributes.

\section{Emotic Dataset}
\label{emotic_dataset}

The EMOTIC dataset is a collection of images of people in unconstrained environments annotated according to their apparent emotional states. The dataset contains $23,571$ images and $34,320$ annotated people. Some of the images were manually collected from the Internet by Google search engine. For that we used a combination of queries containing various places, social environments, different activities and a variety of keywords on emotional states. The rest of images belong to $2$ public benchmark datasets: COCO \cite{mscoco} and Ade20k \cite{ade20k}.  Overall, the images show a wide diversity of contexts, containing people in different places, social settings, and doing different activities.

Fig. \ref{example_annotations} shows three examples of annotated images in the EMOTIC dataset. Images were annotated using Amazon Mechanical Turk (AMT). Annotators were asked to label each image according to what they think  people in the images are feeling. Notice that we have the capacity of making reasonable guesses about other people's emotional state due to our capacity of being empathetic, putting ourselves into another's situation, and also because of our common sense knowledge and our ability for reasoning about visual information. For example, in Fig. \ref{example_annotations}.b, the person is performing an activity that requires \textit{Anticipation} to adapt to the trajectory. Since he is doing a thrilling activity, he seems \textit{excited} about it and he is \textit{engaged} or focused in this activity. In Fig. \ref{example_annotations}.c, the kid feels a strong desire (\textit{yearning}) for eating the chocolate instead of the apple. Because of his situation we can interpret his facial expression as \textit{disquietness} and \textit{annoyance}. Notice that images are also annotated according to the continuous dimensions $Valence$, $Arousal$, and $Dominance$. We describe the emotion annotation modalities of EMOTIC dataset and the annotation process in sections \ref{emo_rep} and \ref{sec_collect_annotations}, respectively.




After the first round of annotations (1 annotator per image), we divided the images into three sets: Training ($70\%$), Validation ($10\%$), and Testing ($20\%$) maintaining a similar affective category distribution across the different sets. After that, Validation and Testing were annotated by $4$ and $2$ extra annotators respectively. As a consequence, images in the Validation set are annotated by a total of $5$ annotators, while images in the Testing set are annotated by $3$ annotators (these numbers can slightly vary for some images since we removed noisy annotations).  

We used the annotations from the Validation to study the consistency of the annotations across different annotators. This study is shown in section \ref{sec_annotationConsistency}. The data statistics and algorithmic analysis on the EMOTIC dataset are detailed in sections \ref{sec_dataStatistics} and \ref{sec_algorithmicAnalysis} respectively.

\subsection{Emotion representation}
\label{emo_rep}

\begin{figure}
	\centering
	\includegraphics[width=0.999\linewidth]{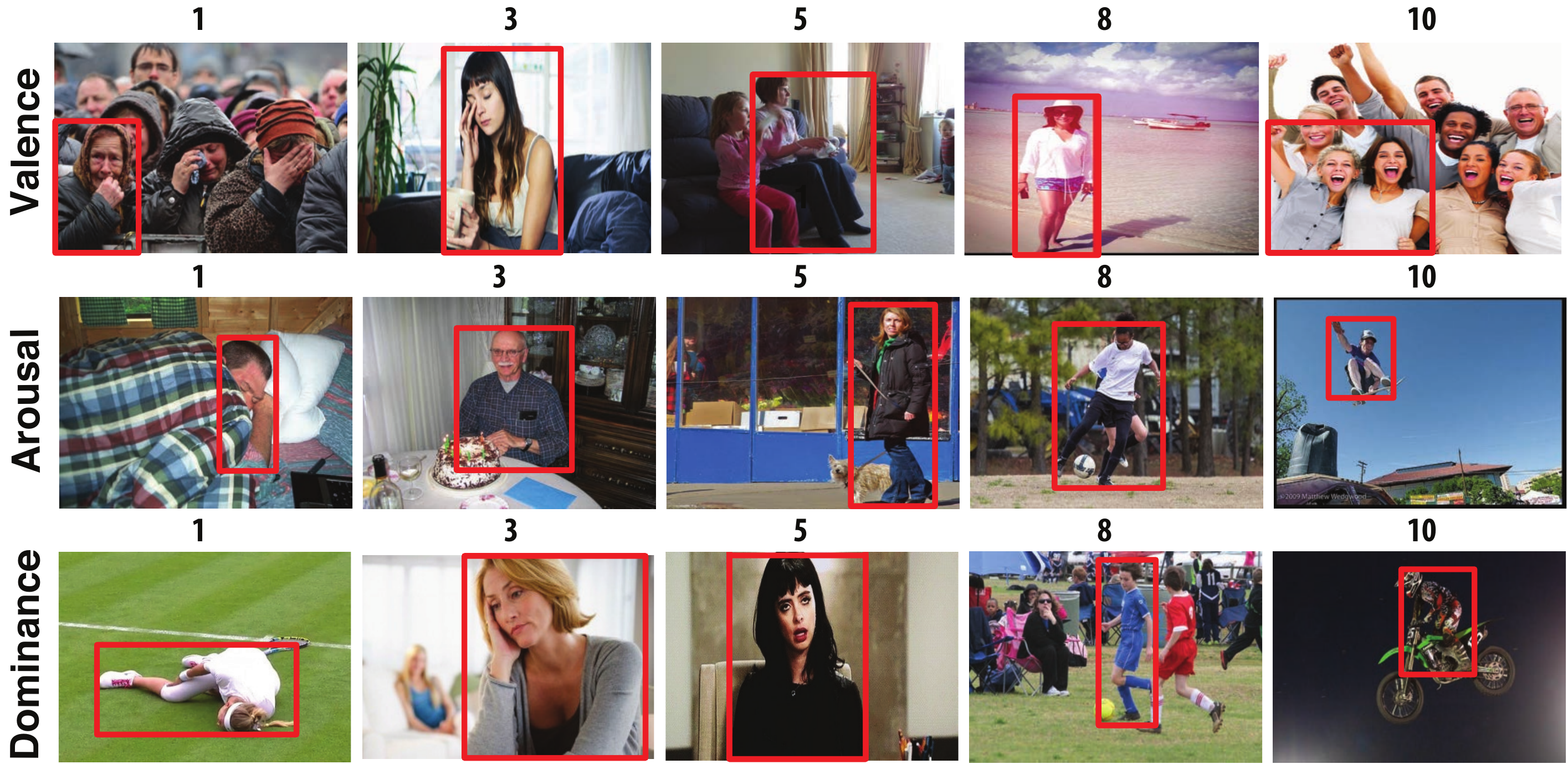}
	\caption{Examples of annotated images in EMOTIC dataset for each of the $3$ continuous dimensions Valence, Arousal \& Dominance. The person in the red bounding box has the corresponding value of the given dimension.}
	\label{sample_images_cont}
\end{figure}

The EMOTIC dataset combines two different types of emotion representation:

\textbf{Continuous Dimensions}: images are annotated according to the $VAD$ model \cite{pad_model}, which represents emotions by a combination of $3$ continuous dimensions: Valence, Arousal and Dominance. In our representation each dimension takes an integer value that lies in the range [$1 - 10$]. Fig. \ref{sample_images_cont} shows examples of people annotated by different values of the given dimension.

\begin{figure*}[!t]
	\centering
	\includegraphics[width=0.999\textwidth]{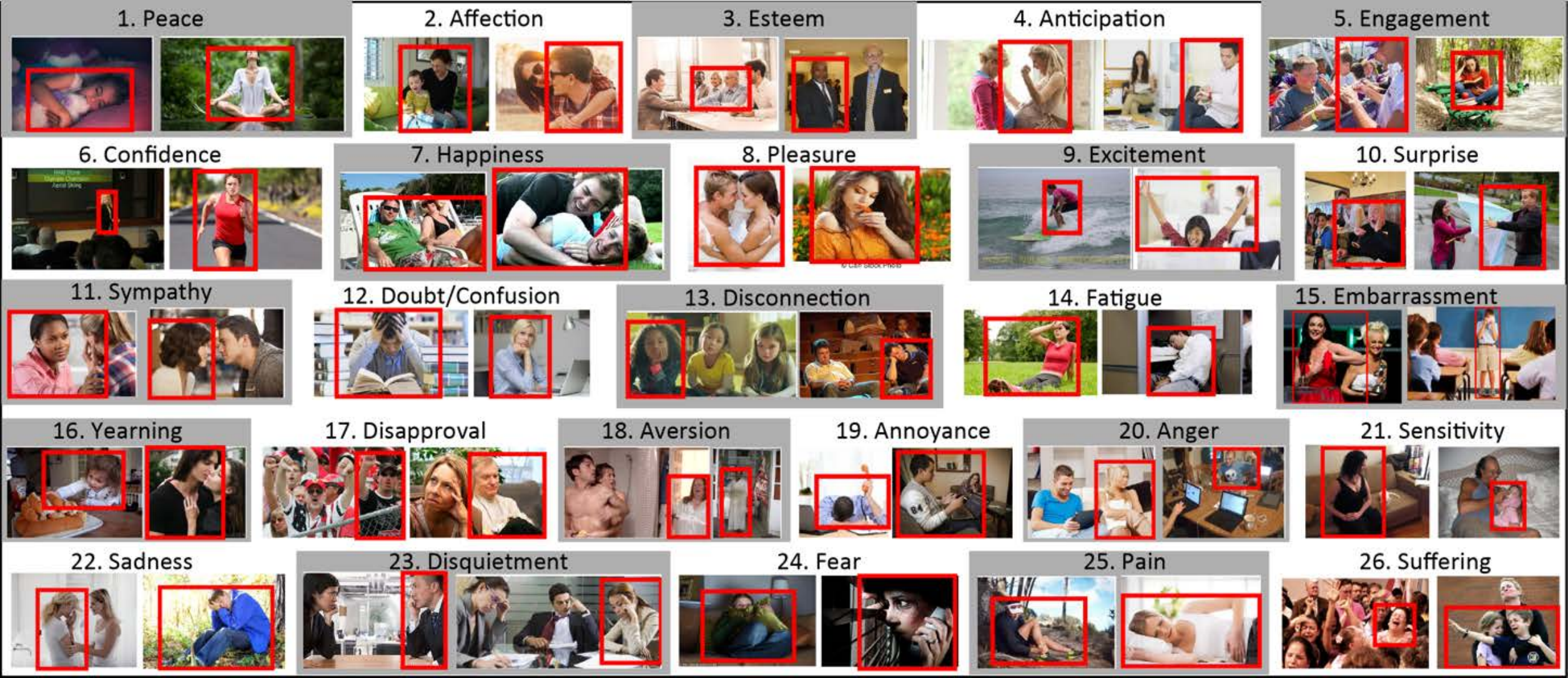}
	\caption{Examples of annotated people in EMOTIC dataset for each of the 26 emotion categories (Table \protect\ref{table_categories}). The person in the red bounding box is annotated by the corresponding category.}
	\label{sample_images_disc}
\end{figure*}

\textbf{Emotion Categories}: in addition to VAD we also established a list of $26$ emotion categories that represent various state of emotions. The list of the $26$ emotional categories and their corresponding definitions can be found in Table \ref{table_categories}. Also, Fig. \ref{sample_images_disc} shows (per category) examples of people showing different emotional categories.

\begin{table} [!t]
	\footnotesize
	\centering
	
	\begin{tabular}{ |p{3.3in}| } 
		
		\hline
		\textbf{1. Affection}: fond feelings; love; tenderness \\
		\hline
		\textbf{2. Anger}: intense displeasure or rage; furious; resentful \\
		\hline
		\textbf{3. Annoyance}: bothered by something or someone; irritated; impatient; frustrated \\
		\hline
		\textbf{4. Anticipation}: state of looking forward; hoping on or getting prepared for possible future events \\
		\hline
		\textbf{5. Aversion}: feeling disgust, dislike, repulsion; feeling hate \\
		\hline
		\textbf{6. Confidence}: feeling of being certain; conviction that an outcome will be favorable; encouraged; proud \\
		\hline
		\textbf{7. Disapproval}: feeling that something is wrong or reprehensible; contempt; hostile \\	
		\hline
		\textbf{8. Disconnection}: feeling not interested in the main event of the surrounding; indifferent; bored; distracted \\
		\hline
		\textbf{9. Disquietment}: nervous; worried; upset; anxious; tense; pressured; alarmed \\
		\hline
		\textbf{10. Doubt/Confusion}: difficulty to understand or decide; thinking about different options \\
		\hline
		\textbf{11. Embarrassment}: feeling ashamed or guilty \\
		\hline
		\textbf{12. Engagement}: paying attention to something; absorbed into something; curious; interested \\
		\hline
		\textbf{13. Esteem}: feelings of favourable opinion or judgement; respect; admiration; gratefulness \\
		\hline
		\textbf{14. Excitement}: feeling enthusiasm; stimulated; energetic \\
		\hline
		\textbf{15. Fatigue}: weariness; tiredness; sleepy \\
		\hline
		\textbf{16. Fear}: feeling suspicious or afraid of danger, threat, evil or pain; horror \\
		\hline
		\textbf{17. Happiness}: feeling delighted; feeling enjoyment or amusement \\
		\hline
		\textbf{18. Pain}: physical suffering \\
		\hline
		\textbf{19. Peace}: well being and relaxed; no worry; having positive \hfil thoughts or sensations; satisfied \\ 
		\hline
		\textbf{20. Pleasure}: feeling of delight in the senses \\
		\hline
		\textbf{21. Sadness}: feeling unhappy, sorrow, disappointed, or discouraged \\
		\hline
		\textbf{22. Sensitivity}: feeling of being physically or emotionally wounded; feeling delicate or vulnerable \\
		\hline
		\textbf{23. Suffering}: psychological or emotional pain; distressed; anguished \\
		\hline
		\textbf{24. Surprise}: sudden discovery of something unexpected \\
		\hline
		\textbf{25. Sympathy}: state of sharing others’ emotions, goals or troubles; supportive; compassionate \\
		\hline
		\textbf{26. Yearning}: strong desire to have something; jealous; envious; lust \\
		\hline
	\end{tabular}
	\caption{Proposed emotion categories with definitions.}
	\vspace{-0.3cm}
    \label{table_categories}
\end{table}



The list of emotion categories has been created as follows. We manually collected an affective vocabulary from dictionaries and books on psychology \cite{oxford,merriam,psy101,affectstates}. This vocabulary consists of a list of approximately $400$ words representing a wide variety of emotional states. After a careful study of the definitions and the similarities amongst these definitions, we formed cluster of words with similar meanings. The clusters were formalized into $26$ categories such that they were distinguishable in a single image of a person with her context. We created the final list of $26$ emotion categories taking into account the \textit{Visual Separability} criterion: words that have a close meaning could not be visually separable. For instance, \textit{Anger} is defined by the words \textit{rage, furious and resentful}. These affective states are different, but it is not always possible to separate them visually in a single image. Thus, our list of affective categories can be seen as a first level of a hierarchy, where each category has associated subcategories. 

Notice that the final list of affective categories also includes the $6$ basic emotions (categories 2, 5, 16, 17, 21, 24), but we used the more general term \textit{Aversion} for the category \textit{Disgust}. Thus, the category \textit{Aversion} also includes the subcategories $dislike$, $repulsion$, and $hate$ apart from \textit{disgust}. 

\subsection{Collecting Annotations}
\label{sec_collect_annotations}
\begin{figure}
	\centering
	\includegraphics[width=0.96\linewidth]{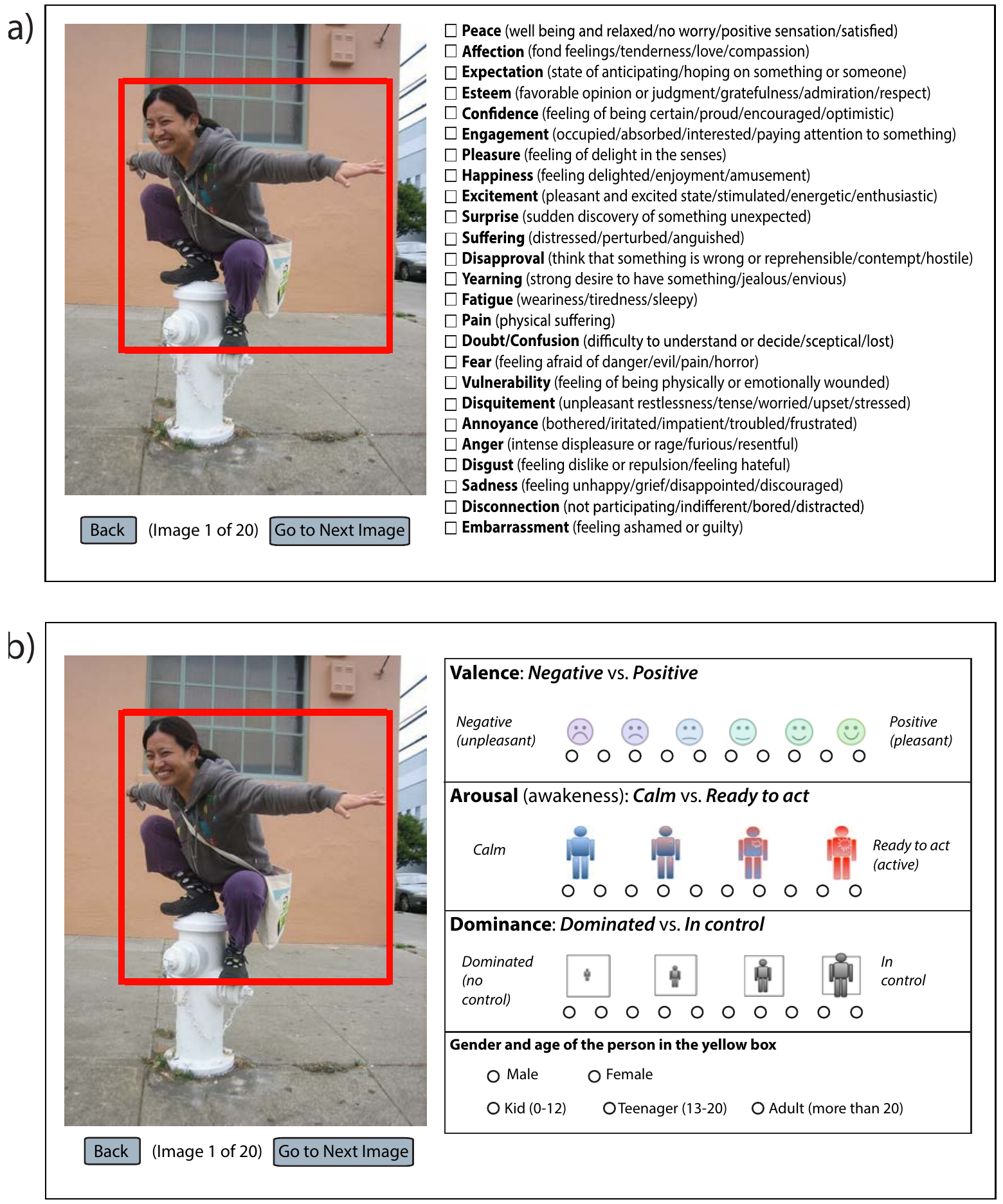}
	\caption{AMT interface designs (a) For Discrete Categories' annotations \& (b) For Continuous Dimensions' annotations}
	\label{interfaces}
\end{figure}

We used Amazon Mechanical Turk (AMT) crowd-sourcing platform to collect the annotations of the EMOTIC dataset. We designed two Human Intelligence Tasks (HITs), one for each of the $2$ formats of emotion representation. The two annotation interfaces are shown in Fig. \ref{interfaces}. Each annotator is shown a person-in-context enclosed in a red bounding-box along with the annotation format next to it. Fig. \ref{interfaces}.a shows the interface for discrete category annotation while Fig. \ref{interfaces}.b displays the interface for continuous dimension annotation. Notice that, in the last box of the continuous dimension interface, we also ask AMT workers to annotate the gender and estimate the age (range) of the person enclosed in red bounding-box. The designing of the annotation interface has two main focuses: \textit{i)} the task is easy to understand and \textit{ii)} the interface fits the HIT in one screen which avoids scrolling.

To make sure annotators understand the task, we showed them how to annotate the images step-wise, by explaining two examples in detail. Also, instructions and examples were attached at the bottom on each page as a quick reference to the annotator. Finally, a summary of the detailed instructions was shown at the top of each page (Table \ref{instructions}).\\

\begin{table}[!t]
	\footnotesize
	\centering
	\begin{tabular}{p{1.6in}|p{1.6in}} 
		\textbf{Emotion Category} & \textbf{Continuous Dimension}\\
		\textit{``Consider each emotion category separately and, if it is applicable to the person in the given context, select that emotion category"}  & \textit{``Consider each emotion dimension separately, observe what level is applicable to the person in the given context, and select that level"}\\
	\end{tabular}
	\caption{Instruction summary for each HIT}
	\label{instructions}
\end{table}

We adopted two strategies to avoid noisy annotations in the EMOTIC dataset. First, we conduct a qualification task to annotator candidates. This qualification task has two parts: (\emph{i}) an Emotional Quotient HIT (based on standard EQ task \cite{eq_test}) and (\emph{ii}) $2$ sample image annotation tasks - one for each of our $2$ emotion representations (discrete categories and continuous dimensions). For the sample annotations, we had a set of acceptable labels. The responses of the annotator candidates to this qualification task were evaluated and those who responded satisfactorily were allowed to annotate the images from the EMOTIC dataset. The second strategy to avoid noisy annotations was to insert, randomly, $2$ control images in every annotation batch of $20$ images; the correct assortment of labels for the control images was know beforehand. Annotators selecting incorrect labels on these control images were not allowed to annotate further and their annotations were discarded. 

\subsection{Agreement Level Among Different Annotators}
\label{sec_annotationConsistency}
 
\begin{figure}[!t]
	\centering
	\includegraphics[width=\linewidth]{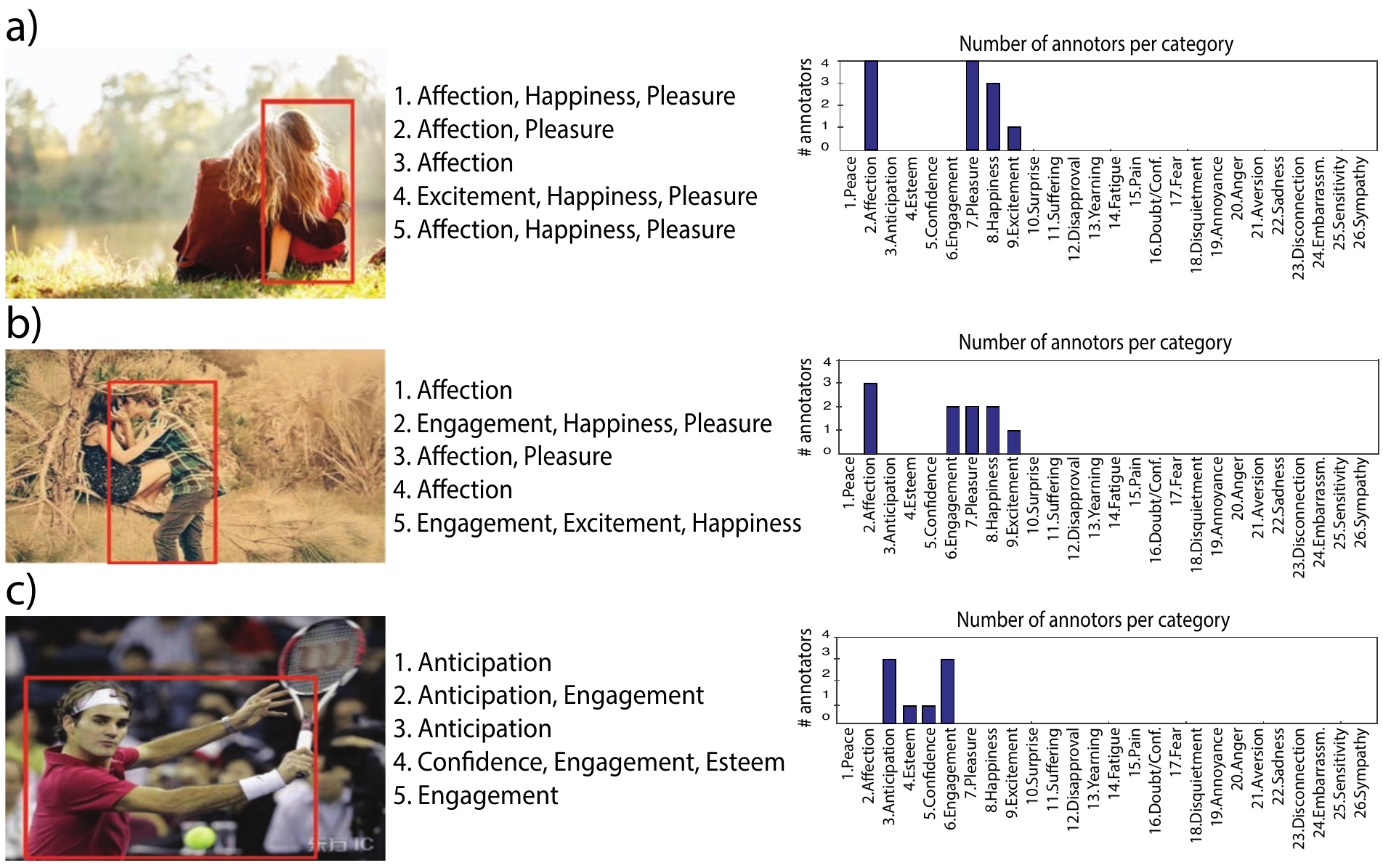}
	\caption{Annotations of five different annotators for 3 images in EMOTIC.}
	\label{multiple_anntrs}
\end{figure}

Since emotion perception is a subjective task, different people can perceive different emotions after seeing the same image. For example in both Fig. \ref{multiple_anntrs}.a and \ref{multiple_anntrs}.b, the person in the red box seems to feel \textit{Affection, Happiness} and \textit{Pleasure} and the annotators have annotated with these categories with consistency. However, not everyone has selected all these emotions. Also, we see that annotators do not agree in the emotions \textit{Excitement} and \textit{Engagement}. We consider, however, that these categories are reasonable in this situation. Another example is that of Roger Federer hitting a tennis ball in Fig. \ref{multiple_anntrs}.c. He is seen predicting the ball (or \textit{Anticipating}) and clearly looks \textit{Engaged} in the activity. He also seems \textit{Confident} in getting the ball. 

\begin{figure}[!t]
	\centering
	\includegraphics[width=\linewidth,height=0.6\linewidth]{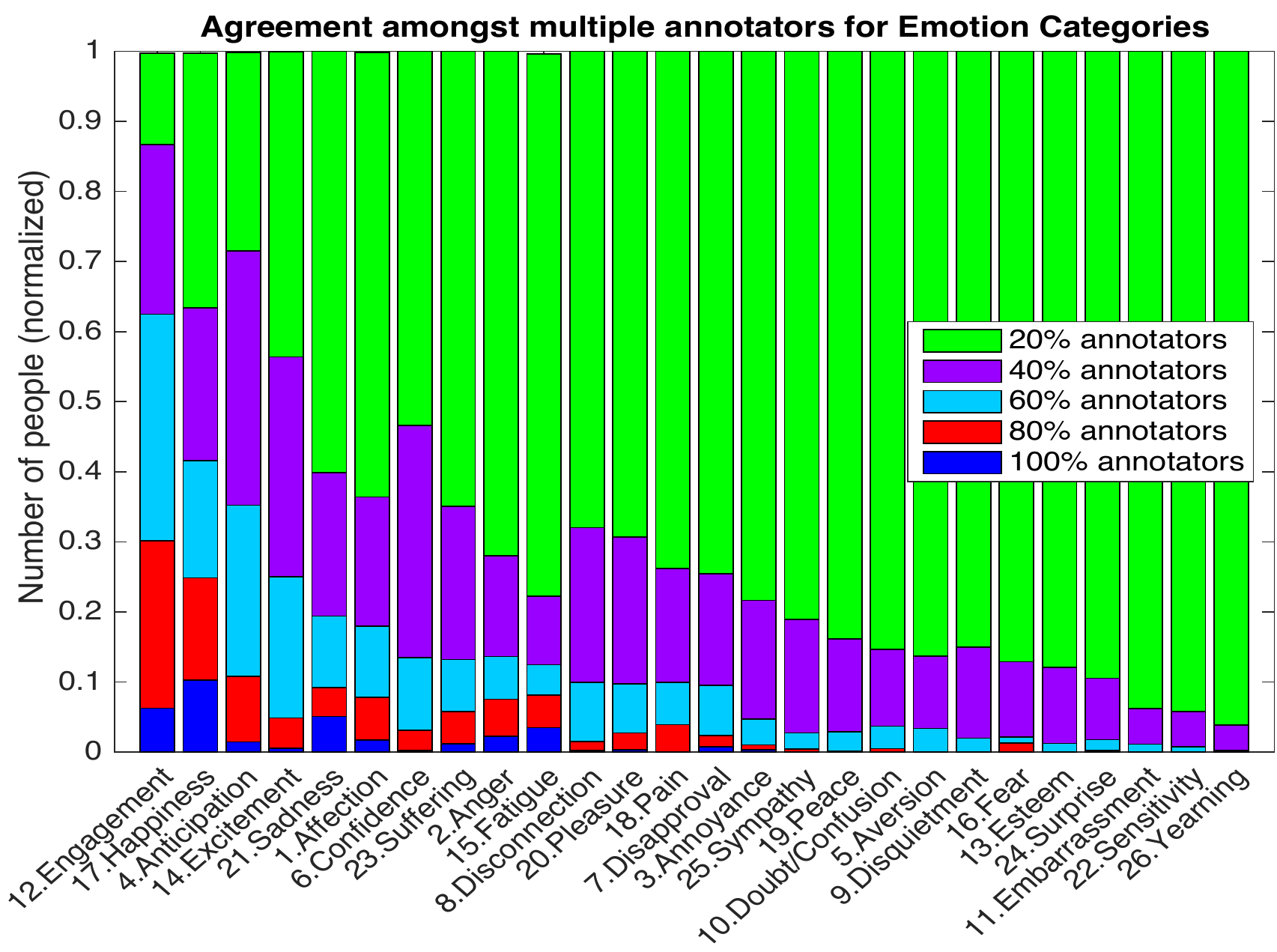}
	\caption{Representation of agreement between multiple annotators. Categories sorted in decreasing order according to the average number of annotators who agreed for that category.}
	\label{agreement_cats}
\end{figure}

After these observations we conducted different quantitative analysis on the annotation agreement. We focused first on analyzing the agreement level in the category annotation. Given a category assigned to a person in an image, we consider as an agreement measure the number of annotators agreeing for that particular category. Accordingly, we calculated, for each category and for each annotation in the validation set, the agreement amongst the annotators and sorted those values across categories. Fig. \ref{agreement_cats} shows the distribution on the percentage of annotators agreeing for an annotated category across the validation set. 

\begin{figure}[!t]
		\includegraphics[width=0.98\linewidth]{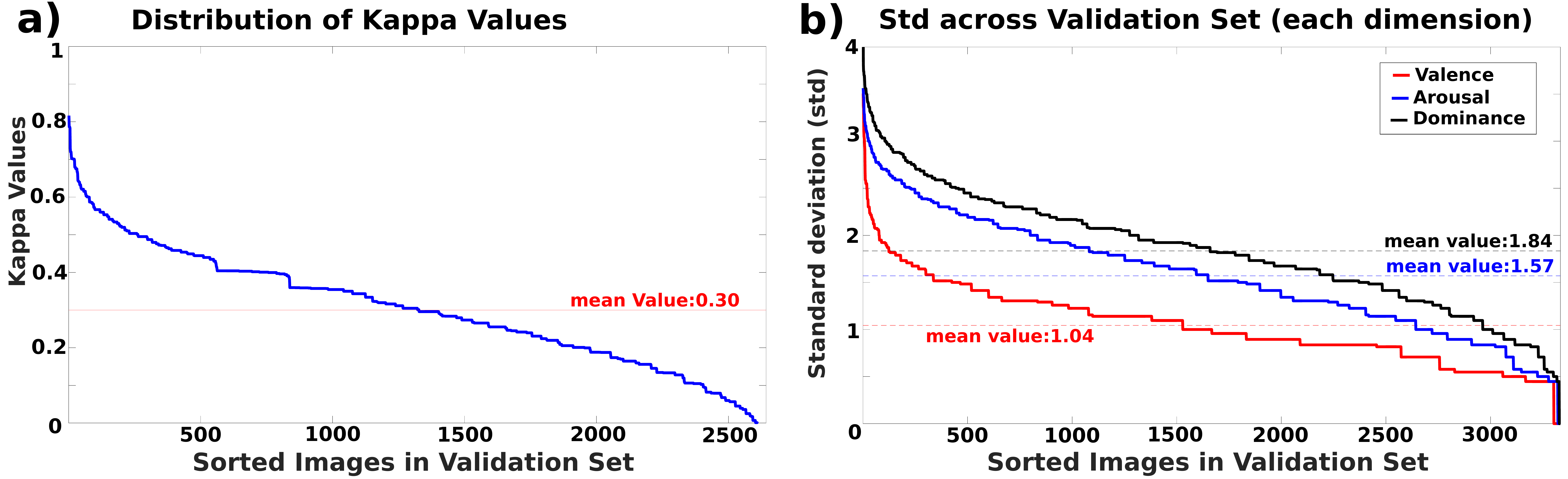}
	\caption{(a) Kappa values (sorted) and (b) Standard deviation (sorted), for each annotated person in validation set}
	\label{kappa_variance}
\end{figure}

We also computed the agreement between all the annotators for a given person using \textit{Fleiss' Kappa ($\kappa$)}. Fleiss' Kappa is a common measure to evaluate the agreement level among a fixed number of annotators when assigning categories to data. In our case, given a person to annotate, there is a subset of $26$ categories. If we have $N$ annotators per image, that means that each of the $26$ categories can be selected by $n$ annotators, where $0\leq n \leq N$. Given an image we compute the Fleiss' Kappa per each emotion category first, and then the general agreement level on this image is computed as the average of these Fleiss' Kappa values across the different emotion categories. We obtained that more than $50\%$ of the images have $\kappa>0.30$. Fig. \ref{kappa_variance}.a shows the distribution of kappa values across the validation set for all the annotated people in the validation set, sorted in decreasing order. Random annotations or total disagreement produces $\kappa\sim0$, however for our case, $\kappa\sim0.3$ (on average) suggesting significant agreement level even though the task of emotion recognition is subjective. 

For continuous dimensions, the agreement is measured by the standard deviation (SD) of the different annotations. The average SD across the Validation set is $1.04$, $1.57$ and $1.84$ for Valence, Arousal and Dominance respectively - indicating that Dominance has higher ($\pm 1.84$) dispersion than the other dimensions. It reflects that annotators disagree more often for Dominance than for the other dimensions which is understandable since Dominance is more difficult to interpret than Valence or Arousal \cite{pad_model}. As a summary, Fig. \ref{kappa_variance}.b shows the standard deviations of all the images in the validation set for all the $3$ dimensions, sorted in decreasing order. 

\subsection{Dataset Statistics}
\label{sec_dataStatistics}

EMOTIC dataset contains $34,320$ annotated people, where $66\%$ of them are males and $34\%$ of them are females. There are $10\%$ children, $7\%$ teenagers and $83\%$ adults amongst them. 

\begin{figure}[!t]
	\centering
	\includegraphics[width=0.90\linewidth]{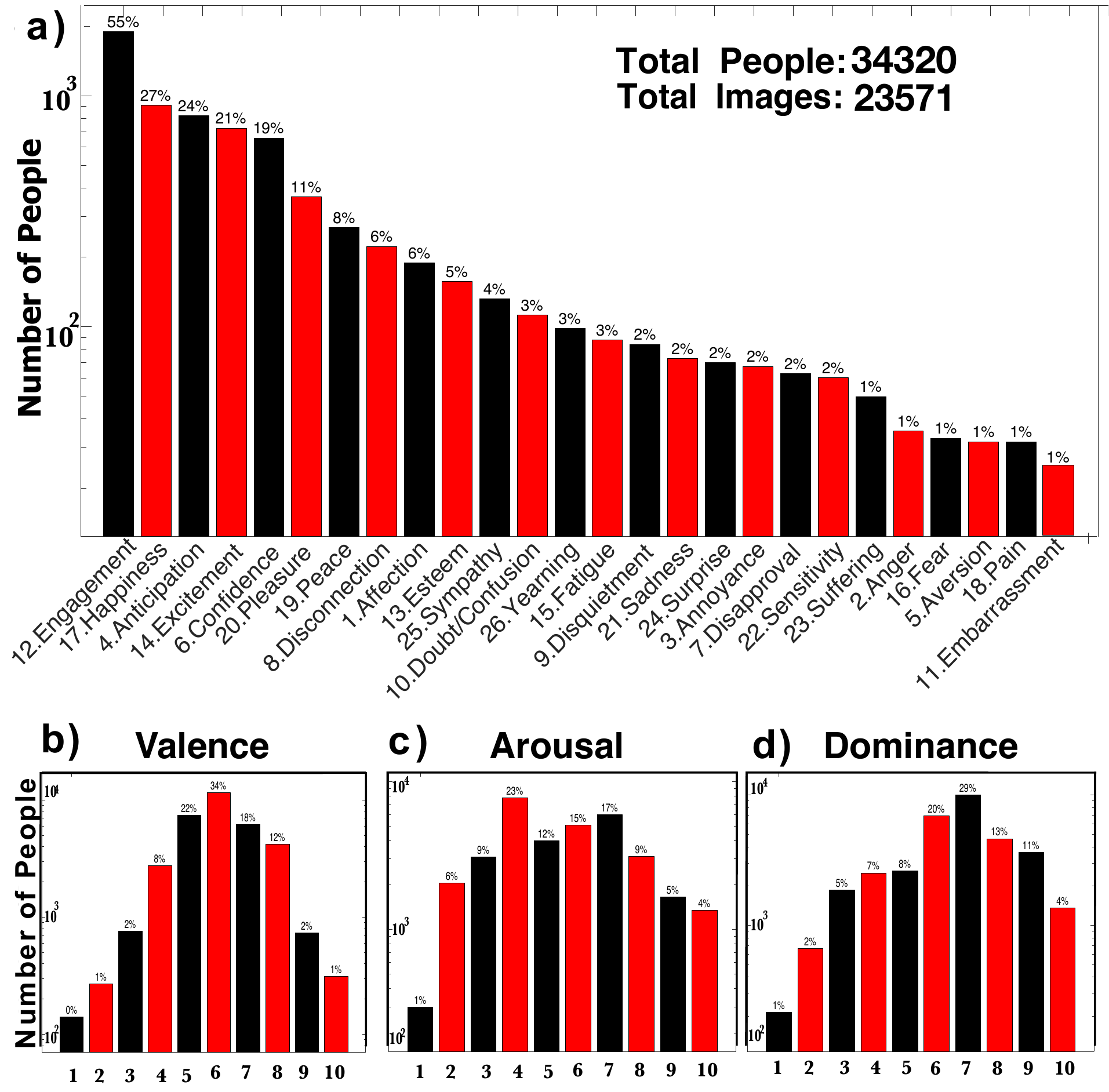}
	\caption{Dataset Statistics. (a) Number of people annotated for each emotion category; (b), (c) \& (d) Number of people annotated for every value of the three continuous dimensions \textit{viz.}Valence, Arousal \& Dominance }
	\label{histogram_annotations}
\end{figure}

Fig. \ref{histogram_annotations}.a shows the number of annotated people for each of the $26$ emotion categories, sorted by decreasing order. Notice that the data is unbalanced, which makes the dataset particularly challenging. An interesting observation is that there are more examples for categories associated to positive emotions, like \emph{Happiness} or \emph{Pleasure}, than for categories associated with negative emotions, like \emph{Pain} or \emph{Embarrassment}. The category with most examples is \emph{Engagement}. This is because in most of the images people are doing something or are involved in some activity, showing some degree of engagement. Figs. \ref{histogram_annotations}.b, \ref{histogram_annotations}.c and \ref{histogram_annotations}.d show the number of annotated people for each value of the $3$ continuous dimensions. In this case we also observe unbalanced data but fairly distributed across the $3$ dimensions which is good for modelling.

\begin{figure}[!t]
	\centering
	\includegraphics[width=\linewidth]{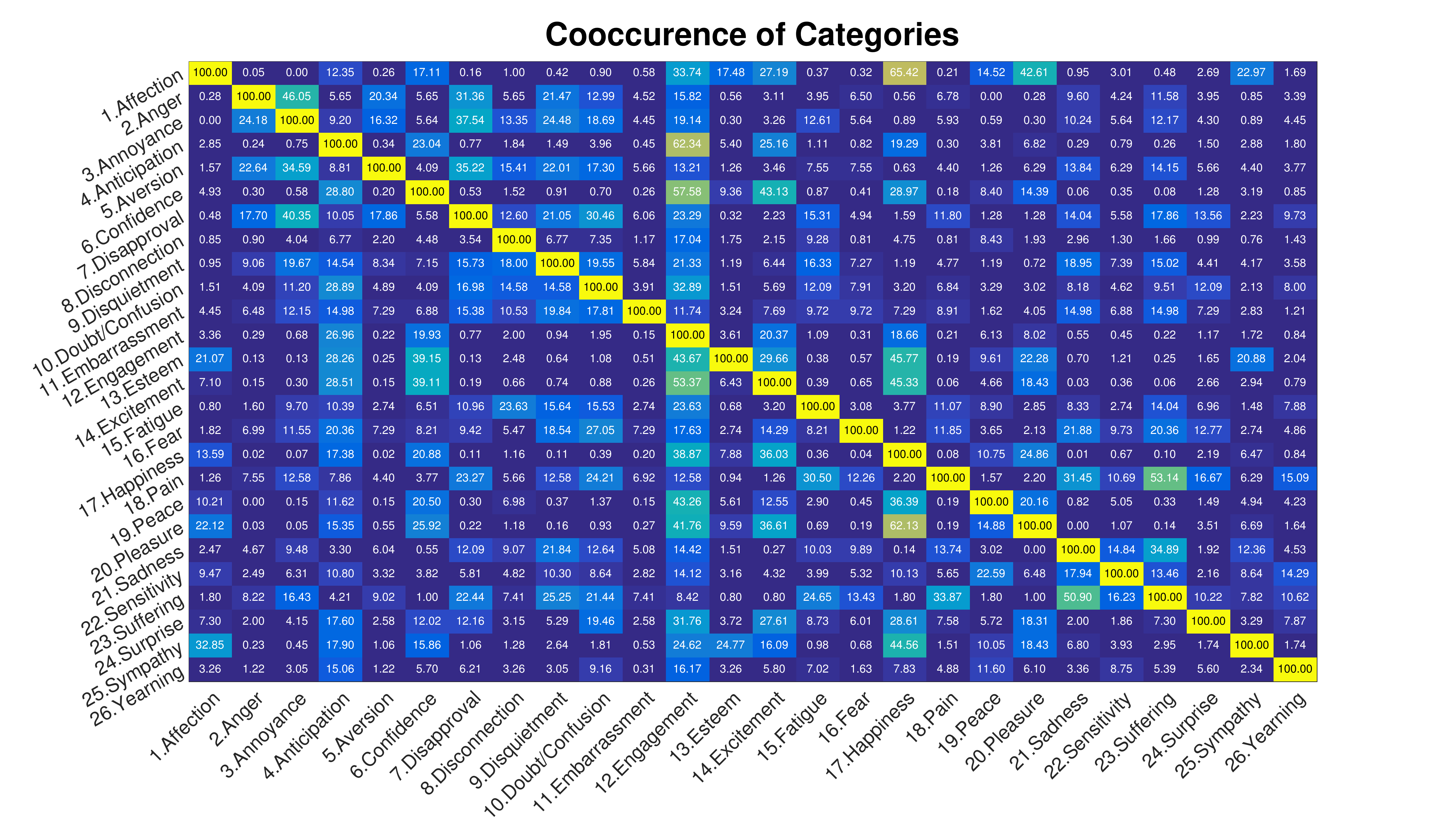}
	\caption{Co-variance between $26$ emotion categories. Each row represents the occurrence probability of every other category given the category of that particular row.}
	\label{covariance_cats}
\end{figure}

Fig. \ref{covariance_cats} shows the co-occurrence rates of any two categories. Every value in the matrix $(r,c)$ ($r$ represents the row category and $c$ column category) is a co-occurrence probability (in \%) of category $r$ if the annotation also contains the category $c$, that is, $P(r|c)$. We observe, for instance, that when a person is labelled with the category \textit{Annoyance}, then there is $46.05\%$ probability that this person is also annotated by the category \textit{Anger}. This means that when a person seems to be feeling \textit{Annoyance} it is likely (by $46.05\%$) that this person might also be feeling \textit{Anger}. We also used a K-Means clustering on the category annotations to find groups of categories that occur frequently. We found, for example, that these category groups are common in the EMOTIC annotations: $\{$\emph{Anticipation}, \emph{Engagement}, \emph{Confidence}$\}$, $\{$\emph{Affection}, \emph{Happiness}, \emph{Pleasure}$\}$, $\{$\emph{Doubt/Confusion}, \emph{Disapproval}, \emph{Annoyance}$\}$, $\{$\emph{Yearning}, \emph{Annoyance}, \emph{Disquietment}$\}$.

\begin{figure}[!t]
	\begin{subfigure}{0.98\linewidth}
		\includegraphics[width=0.98\textwidth,height=0.50\textwidth]{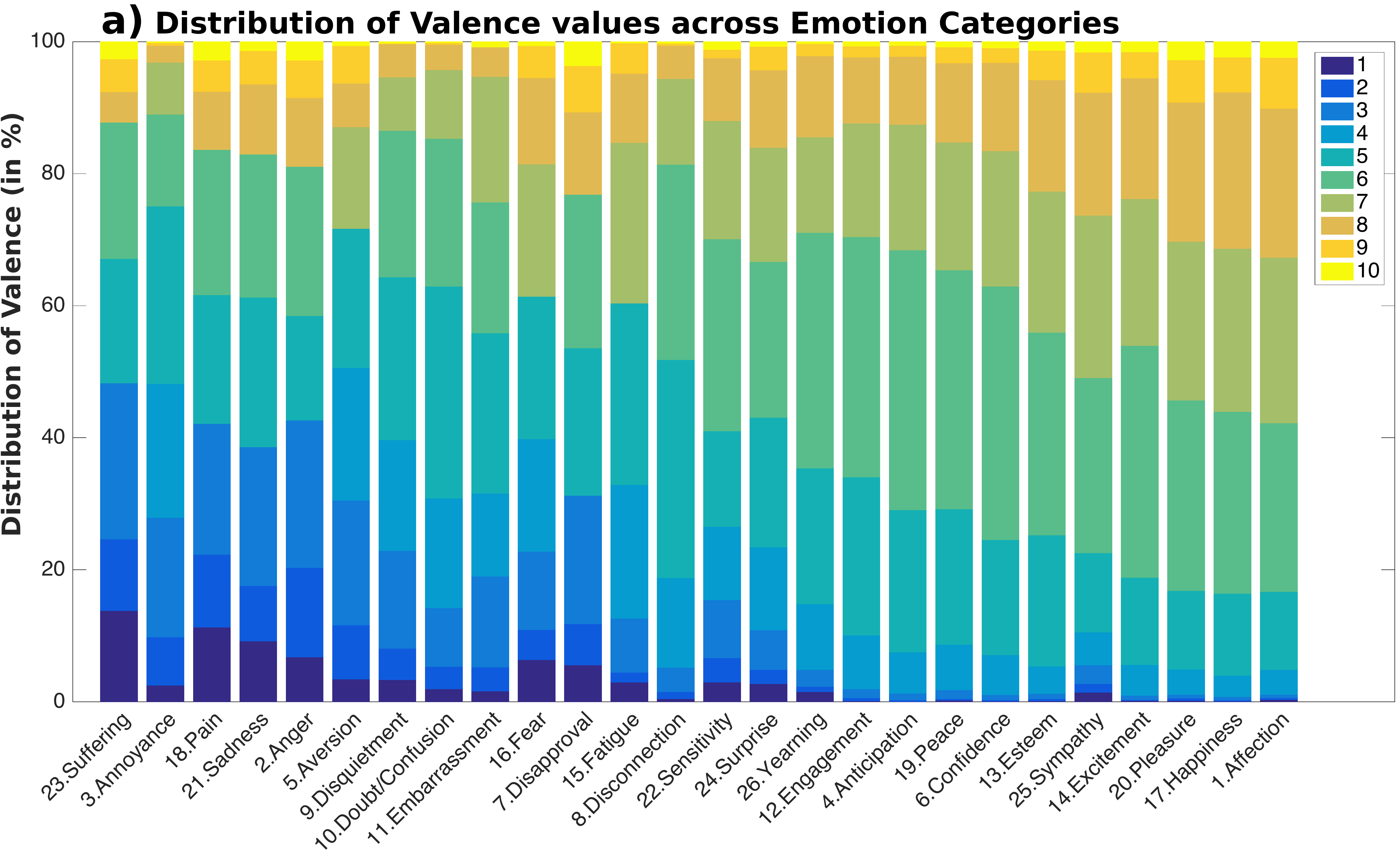}
	\end{subfigure}
	\begin{subfigure}{0.98\linewidth}
		\includegraphics[width=0.98\textwidth,height=0.55\textwidth]{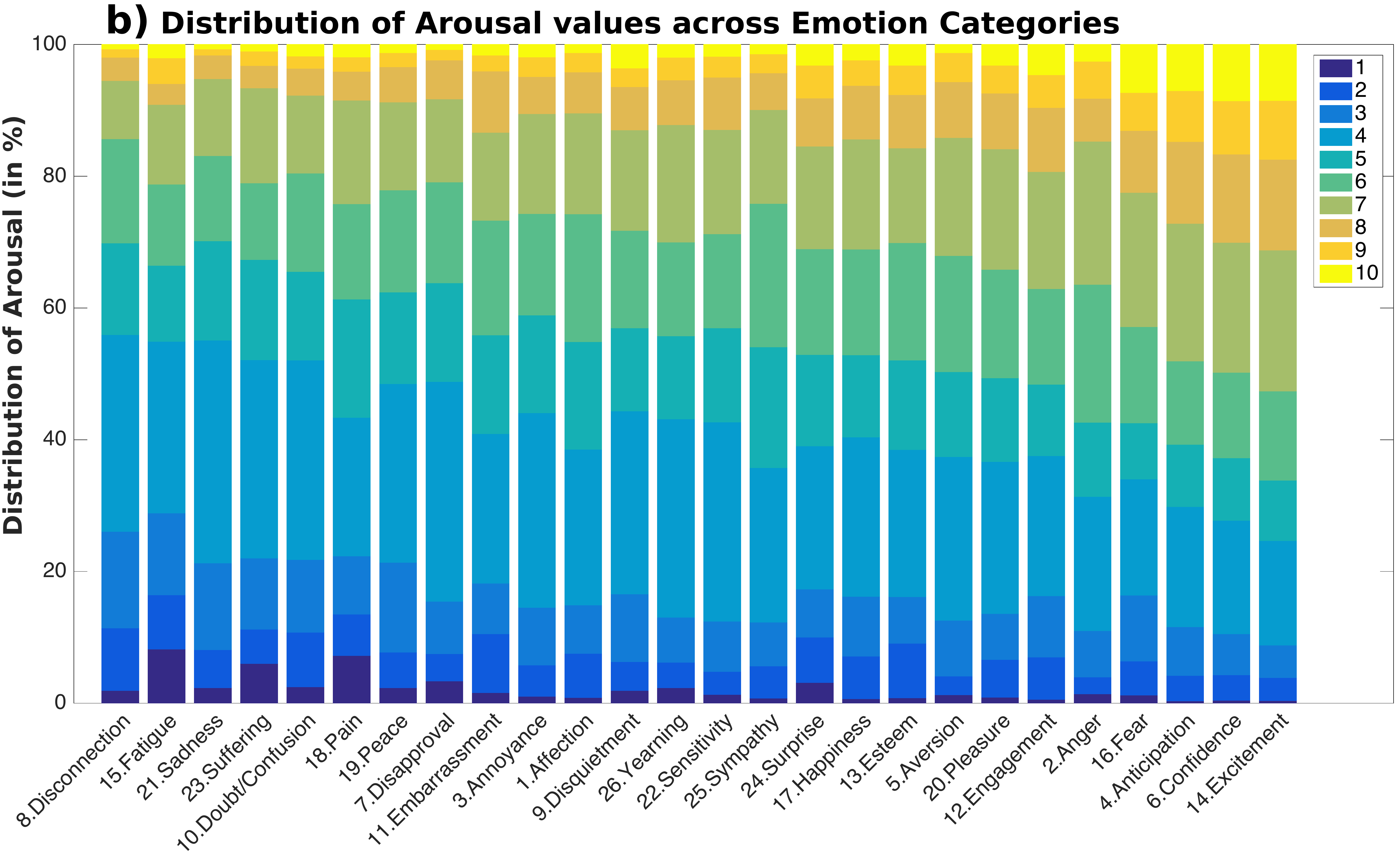}
	\end{subfigure}
	\begin{subfigure}{0.98\linewidth}
		\includegraphics[width=0.98\textwidth,height=0.55\textwidth]{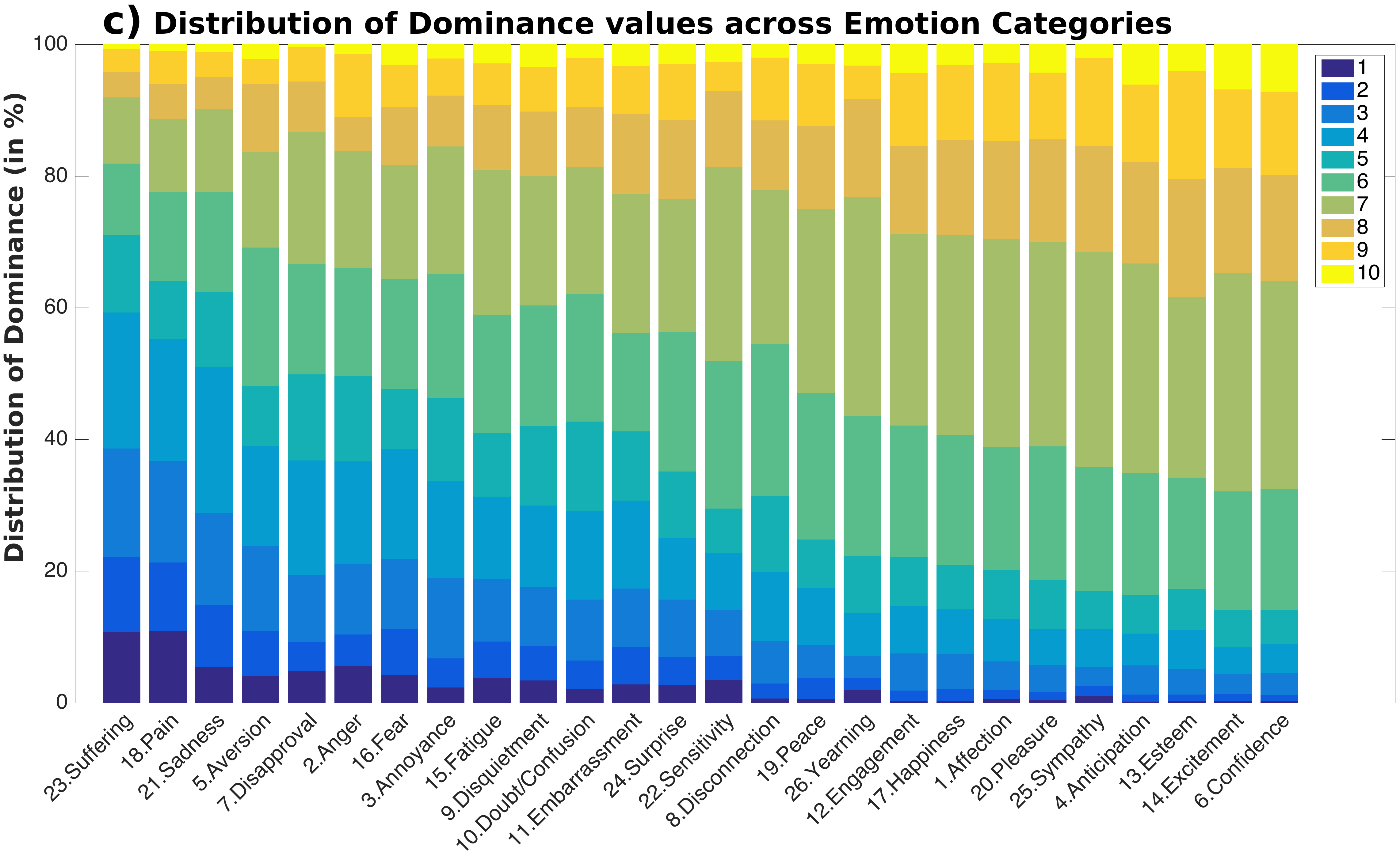}
	\end{subfigure}
	\caption{Distribution of continuous dimension values across emotion categories. Average value of a dimension is calculated for every category and then plotted in increasing order for every distribution.}
	\label{cont_disc}
\end{figure}

Fig. \ref{cont_disc} shows the distribution of each continuous dimension across the different emotion categories. For every plot, categories are arranged in increasing order of their average values of the given dimension (calculated for all the instances containing that particular category). Thus, we observe from Fig. \ref{cont_disc}.a that emotion categories like \textit{Suffering, Annoyance, Pain} correlate with low Valence values (feeling less positive) in average whereas emotion categories like \textit{Pleasure, Happiness, Affection} correlate with higher Valence values (feeling more positive). Also interesting is to note that a category like \textit{Disconnection} lies in the mid-range of Valence value which makes sense. When we observe Fig. \ref{cont_disc}.b, it is easy to understand that emotional categories like \textit{Disconnection, Fatigue, Sadness} show low Arousal values and we see high activeness for emotion categories like \textit{Anticipation, Confidence, Excitement}. Finally, Fig. \ref{cont_disc}.c shows that people are not in control when they show emotion categories like \textit{Suffering, Pain, Sadness} whereas when the Dominance is high, emotion categories like \textit{Esteem, Excitement, Confidence} occur more often. 	

An important remark about the EMOTIC dataset is that there are people whose faces are not visible. More than $25\%$ of the people in EMOTIC have their faces partially occluded or with very low resolution, so we can not rely on facial expression analysis for recognizing their emotional state.

\subsection{Algorithmic Scene Context Analysis}
\label{sec_algorithmicAnalysis}
\begin{figure}[!t]
	\centering
	\includegraphics[width=0.48\textwidth]{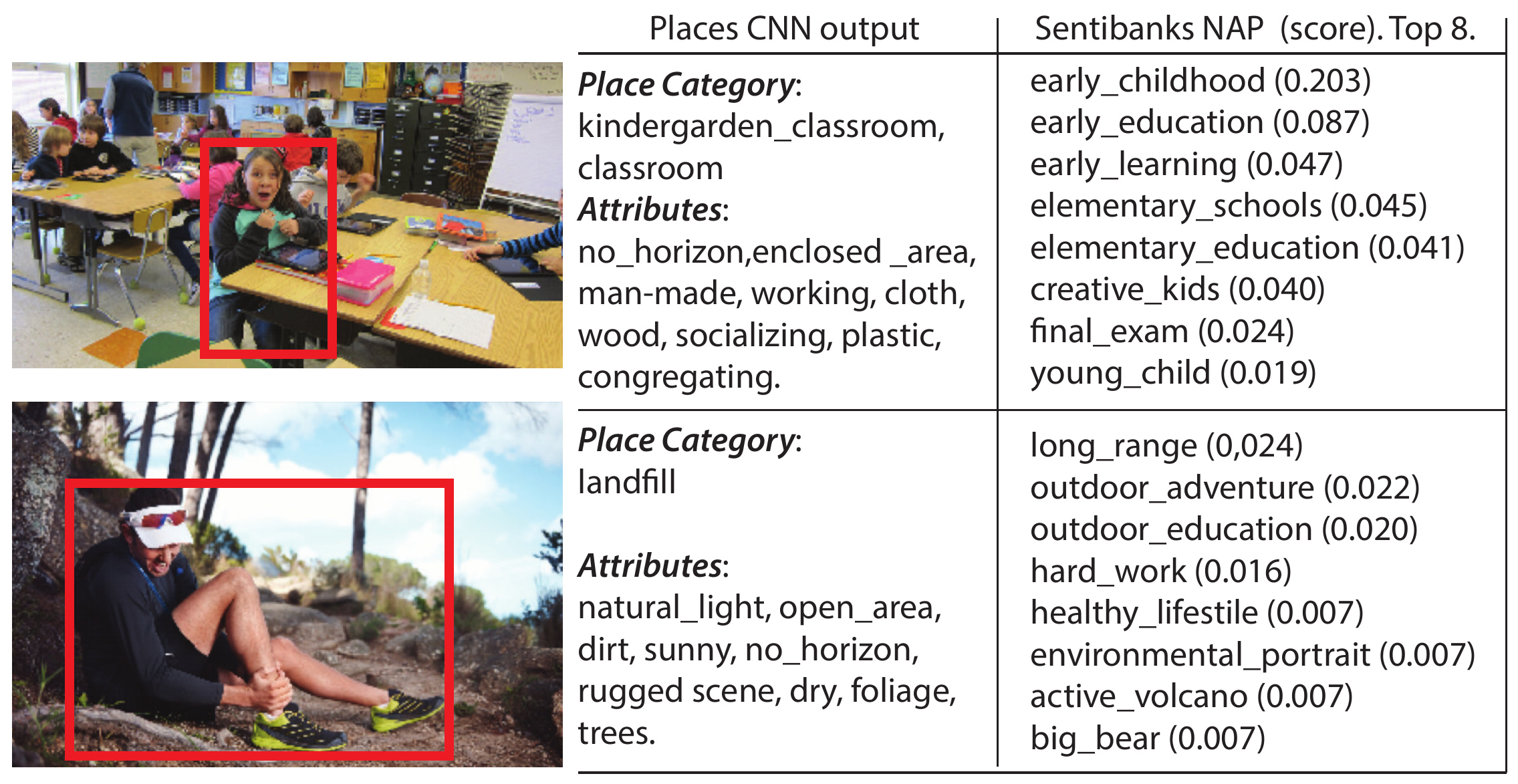}
	\caption{Illustration of $2$ current scene-centric methods for extracting contextual features from the scene: AlexNet Places CNN outputs (place categories and attributes) and Sentibanks ANP outputs for three example images of the EMOTIC dataset.}
	\label{fig_context_info}
\end{figure}

This section illustrates how current scene-centric systems can be used to extract contextual information that can be potentially useful for emotion recognition. In particular, we illustrate this idea with a CNN trained on Places dataset \cite{Places2} and with the Sentibanks Adjective-Noun Pair (ANP) detectors \cite{anp,darrell}, a Visual Sentiment Ontology for Image Sentiment Analysis. As a reference, Fig. \ref{fig_context_info} shows Places and ANP outputs for sample images of the EMOTIC dataset. 

We used AlexNet Places CNN \cite{Places2} to predict the scene category and scene attributes for the images in EMOTIC. This information helps to divide the analysis into place category and place attribute. We observed that the distribution of emotions varies significantly among different place categories. For example, we found that people in the 'ski\_slope' frequently experience \textit{Anticipation} or \textit{Excitement}, which are associated to the activities that usually happen in this place category. Comparing sport-related and working-environment related images, we find that people in sport-related images usually show \textit{Excitement, Anticipation} and \textit{Confidence}, however they show $Sadness$ or $Annoyance$ less frequently. Interestingly,  $Sadness$ and $Annoyance$ appear with higher frequency in working environments. We also observe interesting patterns when correlating continuous dimensions with place attributes and categories. For instance, places where people usually show high Dominance are sport-related places and sport-related attributes. On the contrary, low Dominance is shown in 'jail_cell' or attributes like 'enclosed_area' or 'working', where the freedom of movement is restricted. In Fig. \ref{fig_context_info}, the predictions by Places CNN describe the scene in general, like in the top image there is a girl sitting in a 'kindergarten_classroom' (places category) which usually is situated in enclosed areas with 'no_horizon' (attributes). 

We also find interesting patterns when we compute the correlation between detected ANPs and emotions labelled in the image. For example, in images with people labelled with \textit{Affection}, the most frequent ANP is 'young\_couple', while in images with people labelled with \textit{Excitement} we found frequently the ANPs 'last\_game' and 'playing\_field'. Also, we observe a high correlation between images with \textit{Peace} and ANP like 'old\_couple' and 'domestic\_scenes', and between \textit{Happiness} and the ANPs  'outdoor\_wedding', 'outdoor\_activities', 'happy\_family' or 'happy\_couple'.


Overall, these observations suggest that some common sense knowledge patterns related with emotions and context could be potentially extracted, automatically, from the data. 


\section{CNN Model for Emotion Recognition in Scene Context}\label{emotic_recognition}
\label{model_design}

\begin{figure}
	\includegraphics[width=\linewidth] {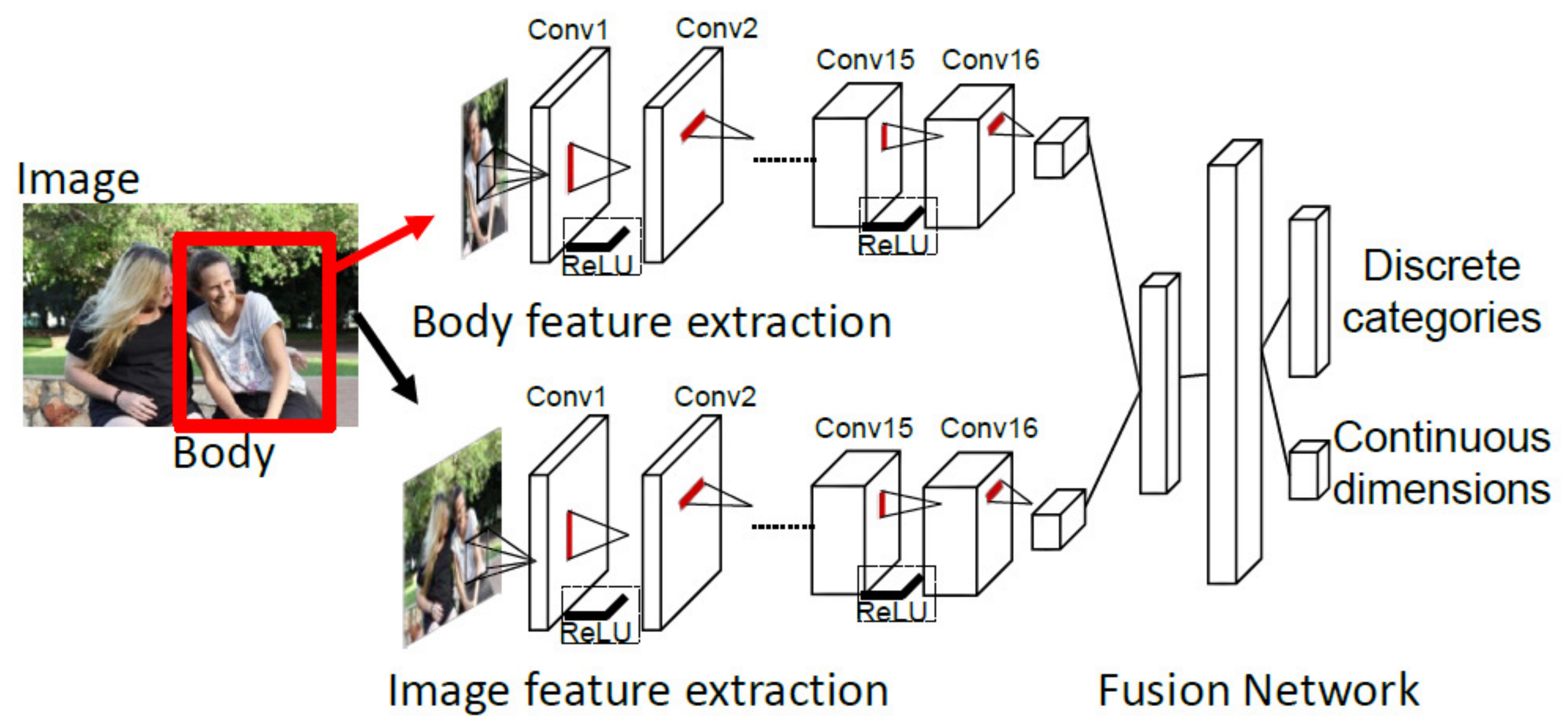}
	\caption{Proposed end-to-end model for emotion recognition in context. The model consists of two feature extraction modules and a fusion network for jointly estimating the discrete categories and the continuous dimensions.}
	\vspace{-0.2cm}
	\label{fig_network_architecture}
\end{figure}

We propose a baseline CNN model for the problem of emotion recognition in context. The pipeline of the model is shown in Fig. \ref{fig_network_architecture} and it is divided in three modules: \textit{body feature extraction}, \textit{image (context) feature extraction} and \textit{fusion network}. The first module takes the whole image as input and generates scene-related features. The second module takes the visible body of the person and generates body-related features. Finally, the third module combines these features to do a fine-grained regression of the two types of emotion representations (section \ref{emo_rep}).

The body feature extraction module takes the visible part of the body of the target person as input and generates body-related features. These features include important cues like face and head aspects and pose or body appearance. In order to capture these aspects, this module is pre-trained with ImageNet \cite{imagenet}, which is an object centric dataset that includes the category \textit{person}. 

The image feature extraction module takes the whole image as input and generates scene-context features. These contextual features can be interpreted as an encoding of the scene category, its attributes and objects present in the scene, or the dynamics between other people present in the scene. To capture these aspects, we pre-train this module with the scene-centric Places dataset \cite{Places2}. 

The fusion module combines features of the two feature extraction modules and estimates the discrete emotion categories and the continuous emotion dimensions.

Both feature extraction modules are based on the one-dimensional filter CNN proposed in \cite{Alvarez16}. These CNN networks provide competitive performance while the number of parameters is low. Each network consists of 16 convolutional layers with 1-dimensional kernels alternating between horizontal and vertical orientations, effectively modeling 8 layers using 2-dimensional kernels. Then, to maintain the location of different parts of the image, we use a global average pooling layer to reduce the features of the last convolutional layer. To avoid internal-covariant-shift we add a batch normalizing layer \cite{batchnorm} after each convolutional layer and rectifier linear units to speed up the training. 

The fusion network module consists of two fully connected (FC) layers. The first FC layer is used to reduce the dimensionality of the features to $256$ and then, a second fully connected layer is used to learn independent representations for each task \cite{RichCaruana}. The output of this second FC layer branches off into $2$ separate representations, one with $26$ units representing the discrete emotion categories, and second with $3$ units representing the 3 continuous dimensions (section \ref{emo_rep}).

\subsection{Loss Function and Training Setup}
\label{training_setup}


We define the loss function as a weighted combination of two separate losses. A prediction $\hat{y}$ is composed by the prediction of each of the $26$ discrete categories and the $3$ continuous dimensions, $\hat{y} = (\hat{y}^{disc},\hat{y}^{cont})$. In particular, $\hat{y}^{disc} = (\hat{y}^{disc}_1,...,\hat{y}^{disc}_{26})$ and $\hat{y}^{cont} = (\hat{y}^{cont}_1,\hat{y}^{cont}_2,\hat{y}^{cont}_{3})$. Given a prediction $\hat{y}$, the loss in this prediction is defined by ${L=\lambda_{disc}L_{disc}+\lambda_{cont}L_{cont}}$, where $L_{disc}$ and $L_{cont}$ represent the loss corresponding to learning the discrete categories and the continuous dimensions respectively. The parameters $\lambda_{(disc,cont)}$ weight the contribution of each loss and are set empirically using the validation set. 

\textbf{Criterion for Discrete categories ($L_{disc}$)}: The discrete category estimation is a multilabel problem with an inherent class imbalance issue, as the number of training examples is not the same for each class (see Fig \ref{histogram_annotations}.a).


In our experiments, we use a weighted Euclidean loss for the discrete categories. Empirically, we found the Euclidean loss to be more effective than using Kullback$-$Leibler divergence or a multi-class multi-classification hinge loss. More precisely, given a prediction $\hat{y}^{disc}$, the weighted Euclidean loss is defined as follows
\begin{equation}
\label{eq_disc_loss}
L_{2disc}(\hat{y}^{disc})=\sum_{i=1}^{26}{w_i}(\hat{y}^{disc}_i - y^{disc}_i)^2
\vspace{-0.02cm}
\end{equation}

\noindent where $\hat{y}^{disc}_i$ is the prediction for the i-th category and $y^{disc}_i$ is the ground-truth label. The parameter $w_i$ is the weight assigned to each category. Weight values are defined as $w_i=\frac{1}{ln(c+p_i)}$, where $p_i$ is the probability of the i-th category and $c$ is a parameter to control the range of valid values for $w_i$. Using this weighting scheme the values of $w_i$ are bounded as the number of instances of a category approach to $0$. This is particularly relevant in our case as we set the weights based on the occurrence of each category for each batch. Experimentally, we obtained better results using this approach compared to setting the global weights based on the entire dataset. 

\textbf{Criterion for Continuous dimensions ($L_{cont}$)}: We model the estimation of the continuous dimensions as a regression problem. Due to multiple annotators annotating the data based on subjective evaluation, we compare the performance when using two different robust losses: (1) a margin Euclidean loss $L_{2cont}$, and (2) the Smooth $L_{1}$ $SL_{1cont}$. The former defines a margin of error ($v_k$) when computing the loss for which the error is not considered. The margin Euclidean loss for continuous dimension is defined as:
\begin{equation}
\label{eq_cont_loss}
L_{2cont}(\hat{y}^{cont})=\sum_{k=1}^{3}{v_k(\hat{y}^{cont}_{k} - y^{cont}_{k})^2},
\vspace{-0.2cm}
\end{equation}
\noindent where $\hat{y}^{cont}_{k}$ and ${y}^{cont}_{k}$ are the prediction and the ground-truth for the k-th dimension, respectively, and $v_k\in\{0,1\}$ is a binary weight to represent the error margin. $v_k=0$ if $|\hat{y}^{cont}_{k} - y^{cont}_{k}| < \theta$. Otherwise, $v_k=1$. If the predictions are within the error margin, \textit{i.e.} error is smaller than $\theta$, then these predictions do not contribute to update the weights of the network. 

The Smooth $L_{1}$ loss refers to the absolute error using the squared error if the error is less than a threshold (set to 1 in our experiments). This loss has been widely used for object detection~\cite{girshick2015fast} and, in our experiments, has been shown to be less sensitive to outliers. Precisely, the Smooth $L_{1}$ loss is defined as follows

\begin{equation}
\label{eq_sl1_cont}
SL_{1cont}(\hat{y}^{cont})=\sum_{k=1}^{3}{v_k}\left\{
                \begin{array}{ll}
                  0.5x^2, if |x_k| < 1\\
                  |x_k| - 0.5, otherwise\\
                \end{array}
              \right.
\vspace{-0.02cm}
\end{equation}
\noindent where $x_k=(\hat{y}_k^{cont} - y_k^{cont})$, and $v_k$ is a weight assigned to each of the continuous dimensions and it is set to $1$ in our experiments. 



We train our recognition system end-to-end, learning the parameters jointly using stochastic gradient descent with momentum. The first two modules are initialized using pre-trained models from Places~\cite{Places2} and Imagenet~\cite{DengDSLL009} while the fusion network is trained from scratch. The batch size is set to $52$ - twice the size of the discrete emotion categories. We found empirically after testing multiple batch sizes (including multiples of $26$ like $26$, $52$, $78$, $108$) that batch-size of $52$ gives the best performance (on the validation set). 

\section{Experiments}\label{experiments}

We trained four different instances of our CNN model, which are the combination of two different input types and the two different continuous loss functions described in section \ref{training_setup}. The input types are body (i.e., upper branch in Fig.~\ref{fig_network_architecture}), denoted by \textbf{B}, and body plus image (i.e., both branches shown in Fig.~\ref{fig_network_architecture}), denoted by \textbf{B+I}. The continuous loss types are denoted in the experiments by $L_2$ for Euclidean loss (equation \ref{eq_cont_loss}) and $SL_1$  for the Smooth $L_1$ (equation \ref{eq_sl1_cont}).

Results for discrete categories in the form of Average Precision per category (the higher, the better) are summarized in Table \ref{comaparison_old_new_data_cat}. Notice that the \textbf{B+I} model outperforms the \textbf{B} model in all categories except $1$. 
The combination of body and image features (\textbf{B+I($SL_1$)} model) is better than the \textbf{B} model.

\begin{table}[!h]
\centering
\begin{tabular}{|l|c|c|c|c|}
\hline
\multirow{2}{*}{{\begin{tabular}[c]{@{}c@{}}\textbf{Emotion} \\ \textbf{Categories}\end{tabular}}} & \multicolumn{4}{c|}{\textbf{CNN Inputs and $L_{cont}$  type}}                                \\ \cline{2-5} 
                                             & \textbf{B ($L_2$)}     & \textbf{B ($SL_1$)} & \textbf{B+I ($L_2$)} & \textbf{B+I ($SL_1$)} \\ \hline
\textbf{1. Affection}                        & 21.80          & 16.55          & 21.16             & \textbf{27.85}    \\ \hline
\textbf{2. Anger}                            & 06.45          & 04.67          & 06.45             & \textbf{09.49}    \\ \hline
\textbf{3. Annoyance}                        & 07.82          & 05.54          & 11.18             & \textbf{14.06}    \\ \hline
\textbf{4. Anticipation}                     & 58.61          & 56.61		   & 58.61             & \textbf{58.64}     \\ \hline
\textbf{5. Aversion}                         & 05.08          & 03.64          & 06.45             & \textbf{07.48}    \\ \hline
\textbf{6. Confidence}                       & 73.79          & 72.57          & 77.97             & \textbf{78.35}    \\ \hline
\textbf{7. Disapproval}                      & 07.63          & 05.50          & 11.00             & \textbf{14.97}    \\ \hline
\textbf{8. Disconnection}                    & 20.78          & 16.12          & 20.37             & \textbf{21.32}    \\ \hline
\textbf{9. Disquietment}                     & 14.32          & 13.99          & 15.54             & \textbf{16.89}    \\ \hline
\textbf{10. Doubt/Confusion}                 & 29.19          & 28.35          & 28.15             & \textbf{29.63}    \\ \hline
\textbf{11. Embarrassment}                   & 02.38          & 02.15          & 02.44             & \textbf{03.18}    \\ \hline
\textbf{12. Engagement}                      & 84.00          & 84.59          & 86.24             & \textbf{87.53}    \\ \hline
\textbf{13. Esteem}                          & 18.36          & \textbf{19.48} & 17.35             & 17.73             \\ \hline
\textbf{14. Excitement}                      & 73.73          & 71.80          & 76.96             & \textbf{77.16}    \\ \hline
\textbf{15. Fatigue}                         & 07.85          & 06.55          & 08.87             & \textbf{09.70}    \\ \hline
\textbf{16. Fear}                            & 12.85          & 12.94          & 12.34             & \textbf{14.14}    \\ \hline
\textbf{17. Happiness}                       & 58.71          & 51.56          & \textbf{60.69}    & 58.26             \\ \hline
\textbf{18. Pain}                            & 03.65          & 02.71          & 04.42             & \textbf{08.94}    \\ \hline
\textbf{19. Peace}                           & 17.85          & 17.09          & 19.43             & \textbf{21.56}    \\ \hline
\textbf{20. Pleasure}                        & 42.58          & 40.98          & 42.12             & \textbf{45.46}    \\ \hline
\textbf{21. Sadness}                         & 08.13          & 06.19          & 10.36             & \textbf{19.66}    \\ \hline
\textbf{22. Sensitivity}                     & 04.23          & 03.60          & 04.82             & \textbf{09.28}    \\ \hline
\textbf{23. Suffering}                       & 04.90          & 04.38          & 07.65             & \textbf{18.84}    \\ \hline
\textbf{24. Surprise}                        & 17.20          & 17.03          & 16.42             & \textbf{18.81}    \\ \hline
\textbf{25. Sympathy}                        & 10.66          & 09.35          & 11.44             & \textbf{14.71}    \\ \hline
\textbf{26. Yearning}                        & 07.82          & 07.40          & \textbf{08.34}    & \textbf{08.34}    \\ \hline \hline
\textbf{Mean}                   & 23.86          & 22.36          & 24.88             & \textbf{27.38}    \\ \hline
\end{tabular}
	\caption{Average Precision (AP) obtained on test set per category. Results for models where the input is just the body \textbf{B}, and models where the input are both the body and the whole image \textbf{B+I}. The type of $L_{cont}$ used is indicated in parenthesis ($L_2$ refers to equation \protect\ref{eq_cont_loss} and $SL_1$ refers to equation \protect\ref{eq_sl1_cont}).} 
	\label{comaparison_old_new_data_cat}%
\end{table}%

\begin{table}[!h]
\centering
\begin{tabular}{|l|c|c|c|c|}
\hline
\multicolumn{1}{|c|}{\multirow{2}{*}{\textbf{\begin{tabular}[c]{@{}c@{}}Continuous \\ Dimensions\end{tabular}}}} & \multicolumn{4}{c|}{\textbf{CNN Inputs and $L_{cont}$ type}}                                    \\ \cline{2-5} 
\multicolumn{1}{|c|}{}                                                                                              & \textbf{B ($L_2$)} & \textbf{B ($SL_1$)} & \textbf{B+I ($L_2$)} & \textbf{B+I ($SL_1$)}  \\ 
\hline
\textbf{Valence}                                                                                                    & 0.0537         & 0.0545             & 0.0546           & 0.0528             \\ \hline
\textbf{Arousal}                                                                                                    & 0.0600         & 0.0630             & 0.0648           & 0.0611             \\ \hline
\textbf{Dominance}                                                                                                  & 0.0570         & 0.0567             & 0.0573           & 0.0579             \\ \hline
\textbf{Mean}                                                                                              & \textbf{0.0569}         & 0.0581    & 0.0589           & 0.0573             \\ \hline
\end{tabular}
\caption{Average Absolute Error (AAE) obtained on test set per each continuous dimension. Results for models where the input is just the body \textbf{B}, and models where the input are both the body and the whole image \textbf{B+I}. The type of $L_{cont}$ used is indicated in parenthesis ($L_2$ refers to equation \protect\ref{eq_cont_loss} and $SL_1$ refers to equation \protect\ref{eq_sl1_cont}).}
	\label{comaparison_old_new_data_con}%
\end{table}

Results for continuous dimensions in the form of Average Absolute Error per dimension, $AAE$ (the lower, the better) are summarized in Table \ref{comaparison_old_new_data_con}. In this case, all the models provide similar results where differences are not significant. 

\begin{figure}[!t]
	\begin{subfigure}{0.5\linewidth}
		\includegraphics[width=0.96\textwidth,height=0.98\textwidth]{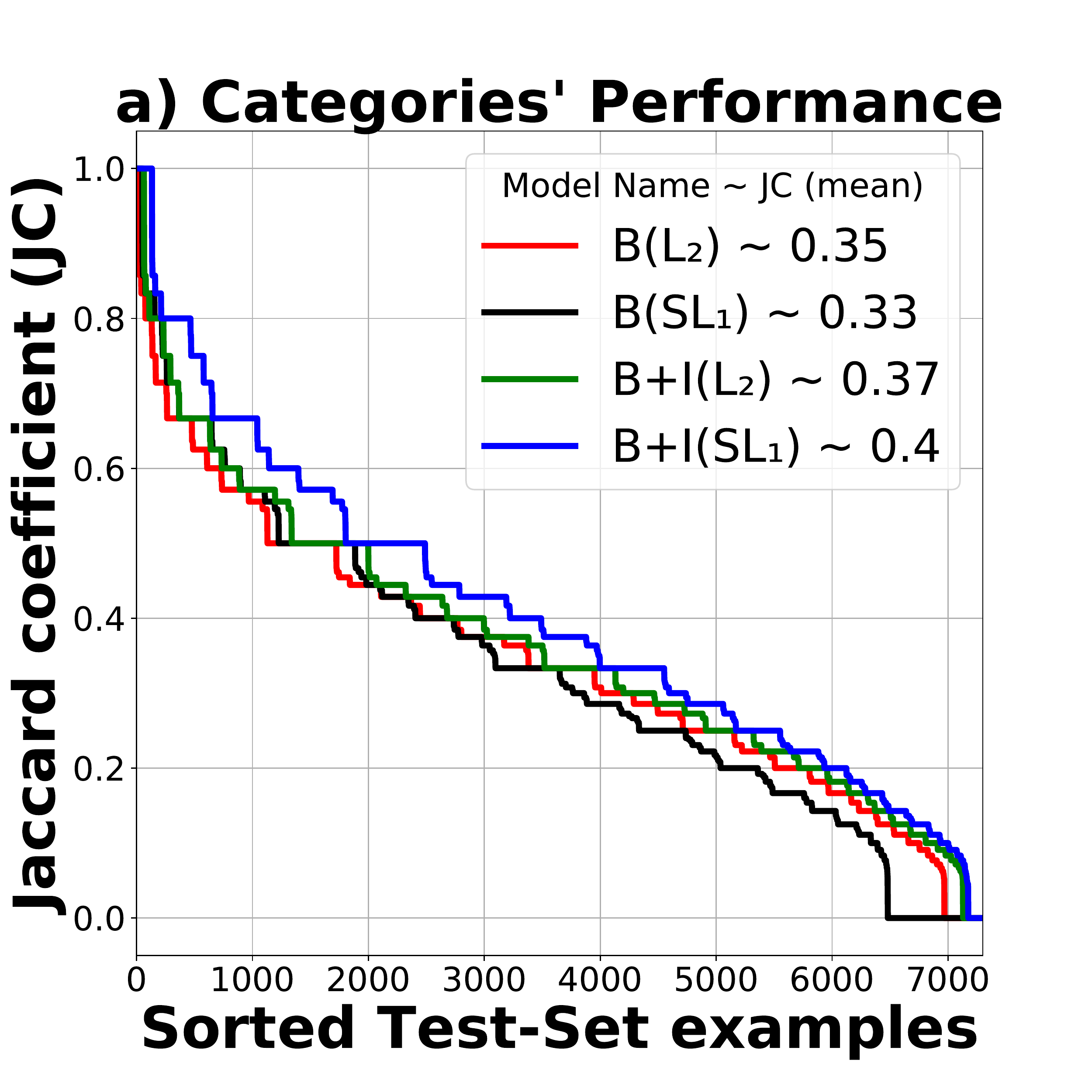}
	\end{subfigure}%
	\begin{subfigure}{0.5\linewidth}
		\includegraphics[width=0.96\textwidth,height=0.98\textwidth]{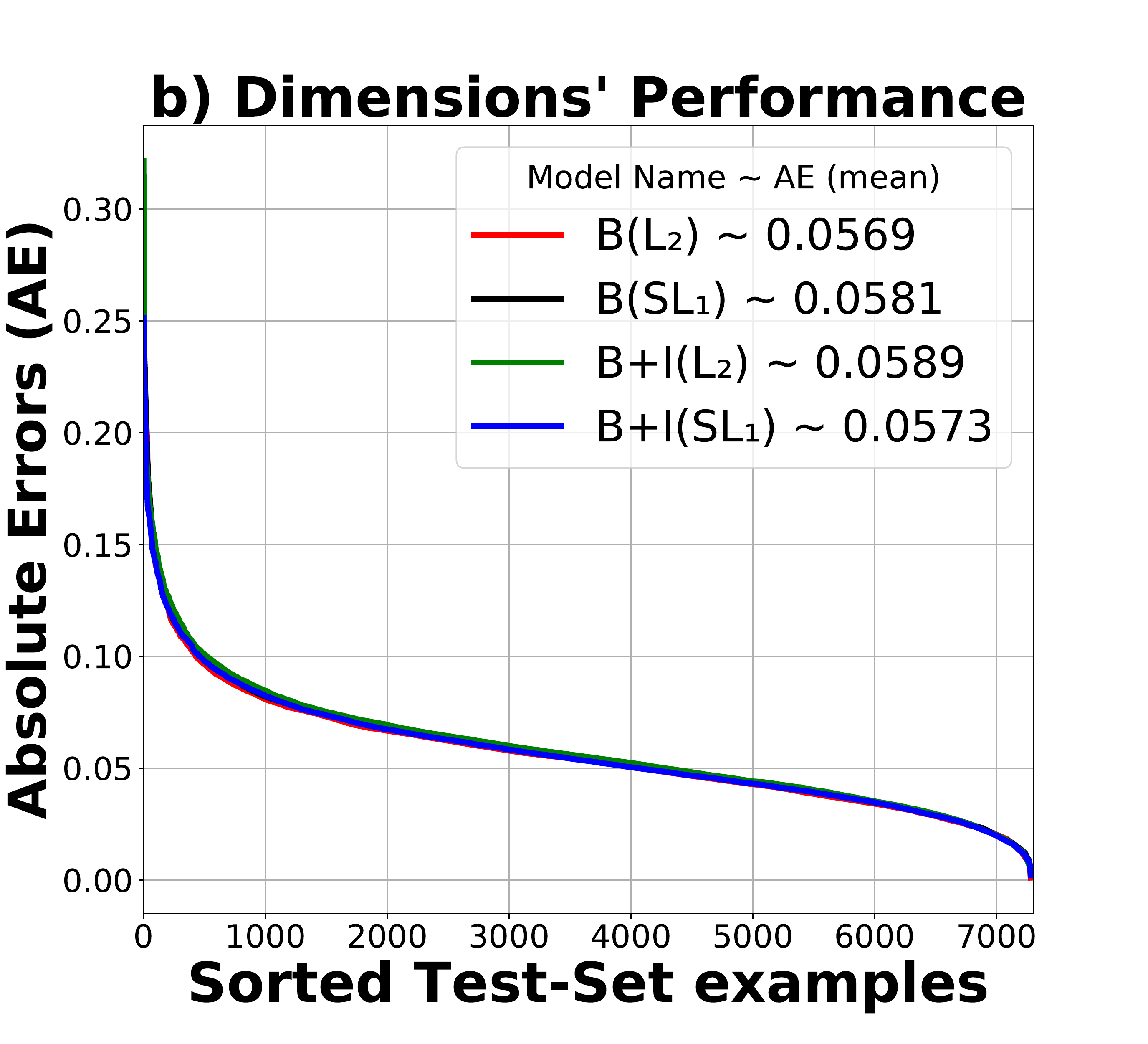}
	\end{subfigure}
	\caption{Results per each sample (Test Set, sorted): (a) Jaccard Coefficient ($JC$) of the recognized discrete categories (b) Average Absolute Error ($AAE$) in the estimation of the three continuous dimensions.}
	\label{cat_con_performance}
	\end{figure}

Fig. \ref{cat_con_performance} shows the summary of the results obtained per each instance in the testing set. Specifically, Fig. \ref{cat_con_performance}.a shows Jaccard coefficient ($JC$) for all the samples in the test set. The $JC$ coefficient is computed as follows: per each category we use as threshold for the detection of the category the value where \textit{Precision $=$ Recall}. Then, the $JC$ coefficient is computed as the number of categories detected that are also present in the ground truth (number of categories in the intersection of detections and ground truth) divided by the total number of categories that are in the ground truth or detected (union over detected categories and categories in the ground truth). The higher this $JC$ is the better, with a maximum value of $1$, where the detected categories and the ground truth categories are exactly the same. In the graphic, examples are sorted in decreasing order of the $JC$ coefficient. Notice that these results also support that the \textbf{B+I} model outperforms the \textbf{B} model. 

For the case of continuous dimensions, Fig.~\ref{cat_con_performance}.b shows the Average Absolute Error ($AAE$) obtained per each sample in the testing set. Samples are sorted by increasing order (best performances on the left). Consistent with the results shown in Table \ref{comaparison_old_new_data_con}, we do not observe a significant  difference among the different models.

\begin{figure*}[!t]
	\centering
	\includegraphics[width=0.999\textwidth]{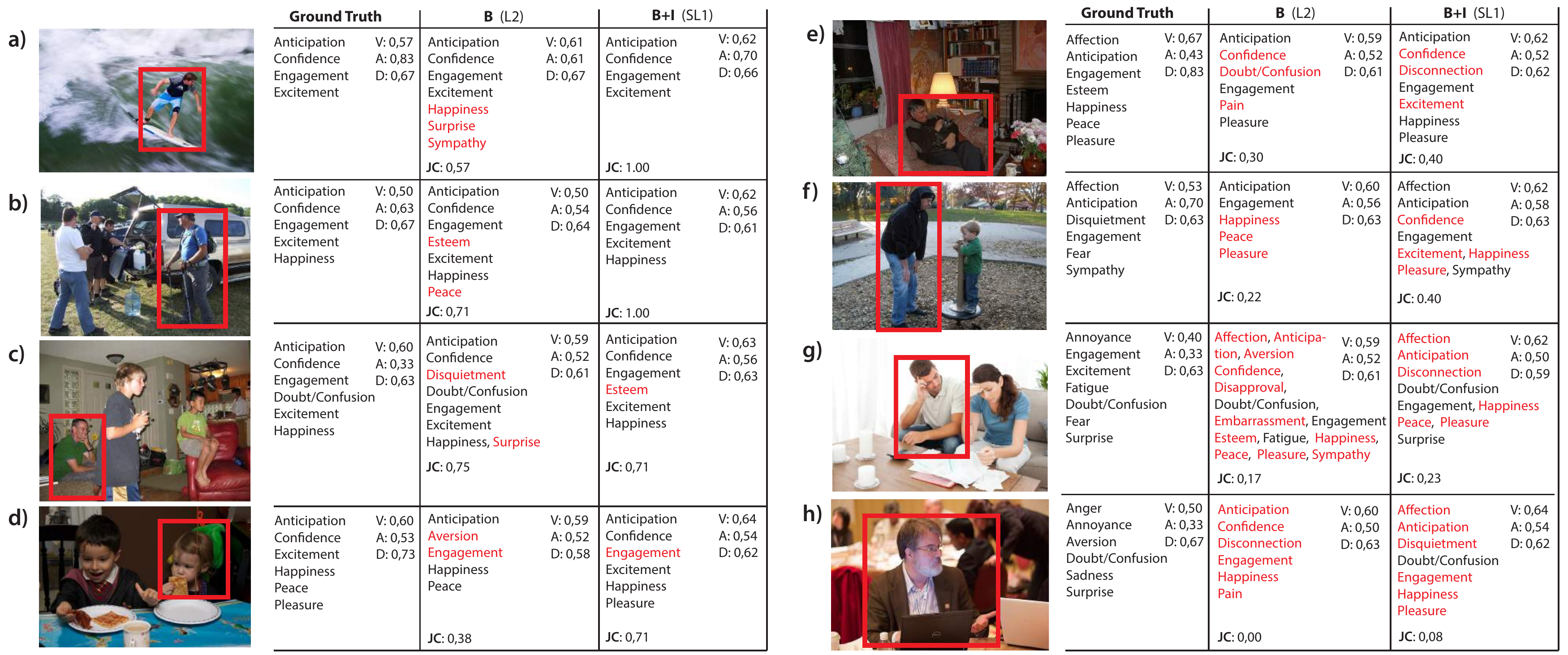}
	\caption{Ground truth and results on images randomly selected with different $JC$ scores.}
	\label{fig_qualitativeResults}
\end{figure*}

Finally, Fig. \ref{fig_qualitativeResults} shows qualitative predictions for the best \textbf{B} and \textbf{B+I} models. These examples were randomly selected among samples with high $JC$ in \textbf{B+I} (a-b) and samples with low $JC$ in \textbf{B+I} (g-h). Incorrect category recognition is indicated in red. As shown, in general, \textbf{B+I} model outperforms \textbf{B}, although there are some exceptions, like Fig. \ref{fig_qualitativeResults}.c. 

\subsection{Context Features Comparison}


The goal of this section is to compare different context features for the problem of emotion recognition in context. A key aspect for incorporating the context in an emotion recognition model is to be able to obtain information from the context that is actually relevant for emotion recognition. Since the context information extraction is a scene-centric task, the information extracted from the context should be based in a scene-centric feature extraction system. That is why our baseline model uses a Places CNN for the context feature extraction module. However, recent works in sentiment analysis (detecting the emotion of a person when he/she observes an image) also provide a system for scene feature extraction that can be used for encoding the relevant contextual information for emotion recognition. 

To compute body features, denoted by \textbf{B$_f$}, we fine tune an AlexNet ImageNet CNN with EMOTIC database, and use the average pooling of the last convolutional layer as features. For the context (image), we compare two different feature types, which are denoted by \textbf{I$_{f}$} and \textbf{I$_{S}$}. \textbf{I$_{f}$} are obtained by fine tunning an AlexNet Places CNN with EMOTIC database, and taking the average pooling of the last convolutional layer as features (similar to \textbf{B$_f$}), while \textbf{I$_{S}$} is a feature vector composed of the sentiment scores for the ANP detectors from the implementation of \cite{darrell}. 

To fairly compare the contribution of the different context features, we train Logistic Regressors for the following features and combination of features: (1) \textbf{B$_{f}$}, (2) \textbf{B$_{f}$}+\textbf{I$_{f}$}, and (3) \textbf{B$_{f}$}+\textbf{I$_{S}$}. For the discrete categories we obtain mean APs $AP=23.00$, $AP=27.70$, and $AP=29.45$, respectively. For the continuous dimensions, we obtain AAE $0.0704$, $0.0643$, and $0.0713$ respectively. We observe that, for the discrete categories, both \textbf{I$_{f}$} and \textbf{I$_{S}$} contribute relevant information to the emotion recognition in context. Interestingly, \textbf{I$_{S}$} performs better than \textbf{I$_{f}$}, even though these features have not been trained using EMOTIC. However, these features are smartly designed for sentiment analysis, which is a problem closely related to extracting relevant contextual information for emotion recognition, and are trained with a large dataset of images.  

\section{Conclusions}
In this paper we pointed out the importance of considering the person scene context in the problem of automatic emotion recognition in the wild. We presented the EMOTIC database, a dataset of $23,571$ natural unconstrained images with $34,320$ people labeled according to their apparent emotions. The images in the dataset are annotated using two different emotion representations: $26$ discrete categories, and the $3$ continuous dimensions $Valence$, $Arousal$ and $Dominance$. We described in depth the annotation process and analyzed the annotation consistency of different annotators. We also provided different statistics and algorithmic analysis on the data, showing the characteristics of the EMOTIC database. In addition, we proposed a baseline CNN model for emotion recognition in scene context that combines the information of the person (body bounding box) with the scene context information (whole image). We also compare two different feature types for encoding the contextual information. Our results show the relevance of using contextual information to recognize emotions and, in conjunction with the EMOTIC dataset, motivate further research in this direction. All the data and trained models are publicly available for the research community in the website of the project.

\appendices
\section*{Acknowledgment}
	This work has been partially supported by the \emph{Ministerio de Economia, Industria y Competitividad (Spain)}, under the Grants Ref. TIN2015-66951-C2-2-R and RTI2018-095232-B-C22, and by Innovation and Universities (FEDER funds). The authors also thank NVIDIA for their generous hardware donations.


\bibliographystyle{IEEEtran}
\bibliography{tpami}

\vfill\eject

\begin{IEEEbiography}[{\includegraphics[width=1in,height=1.35in,clip,keepaspectratio]{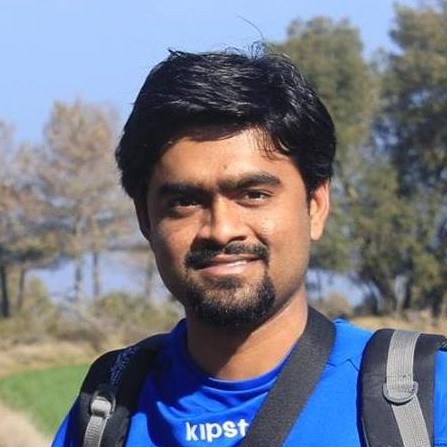}}] {Ronak Kosti}
	is pursuing his PhD at Universitat Oberta de Catalunya, Spain advised by Prof. Agata Lapedriza. He is working with Scene Understanding and Artificial Intelligence (SUNAI) group in computer vision, specifically in the area of affective computing. He obtained his Masters in Machine Intelligence from DA-IICT (Dhirubhai Ambani Institute of Information and Communication Technology) in 2014. His master's research was based on depth estimation from single image using Artificial Neural Networks.
\end{IEEEbiography}

\begin{IEEEbiography}[{\includegraphics[width=1in,height=1.25in,clip,keepaspectratio]{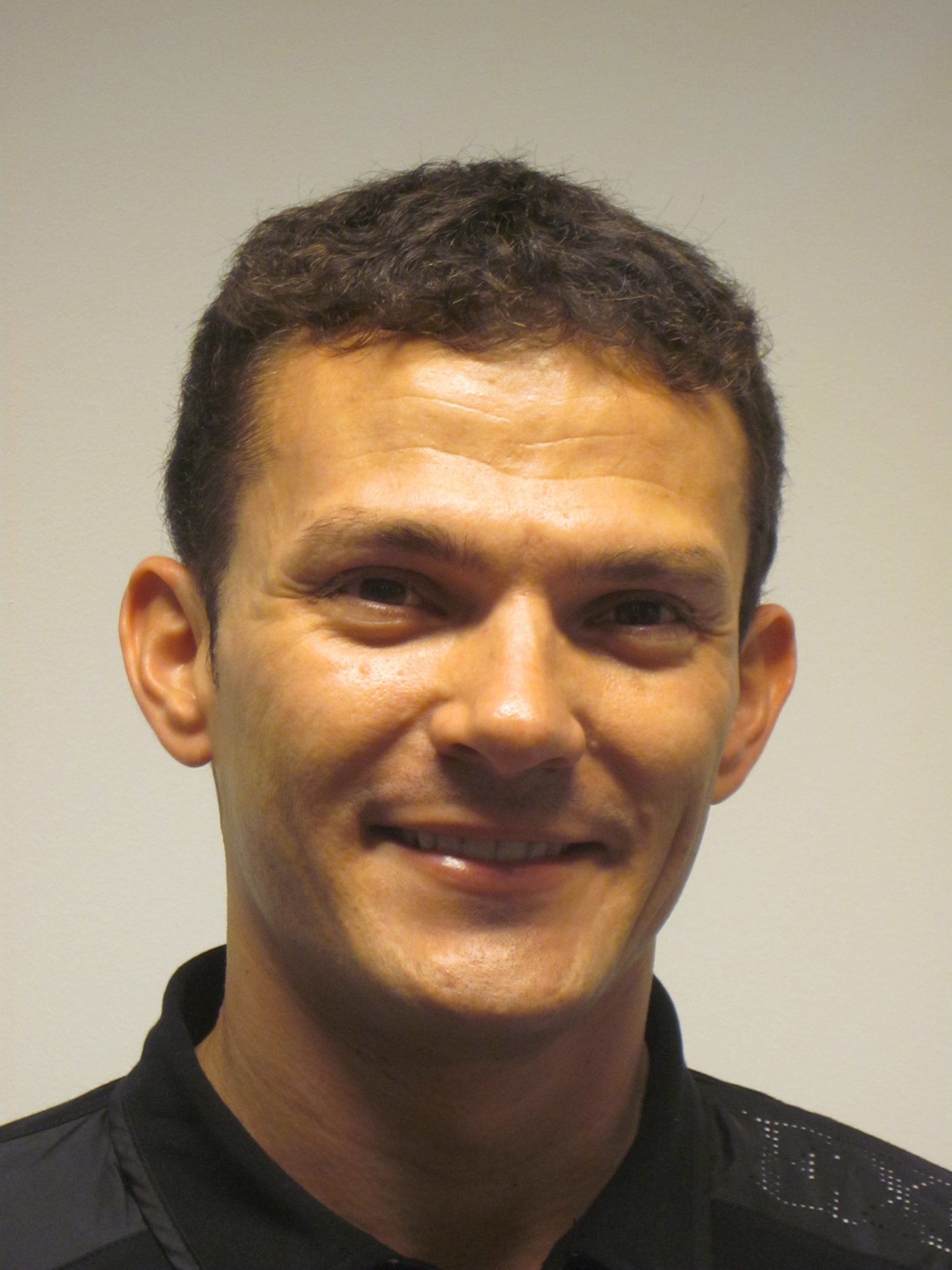}}]{Jose M. Alvarez} is a senior research scientist at NVIDIA. Previously, he was senior deep learning researcher at Toyota Research Institute, US; prior to that he was with Data61, CSIRO, Australia (formerly NICTA) as a researcher. He obtained his Ph.D. in 2010 from Autonomous University of Barcelona under the supervision of Prof. Antonio Lopez and Prof. Theo Gevers. Previous to CSIRO, he worked as a postdoctoral researcher at the Courant Institute of Mathematical Science at New York University under the supervision of Prof. Yann LeCun.
\end{IEEEbiography}

\begin{IEEEbiography}[{\includegraphics[width=1in,height=1.25in,clip,keepaspectratio]{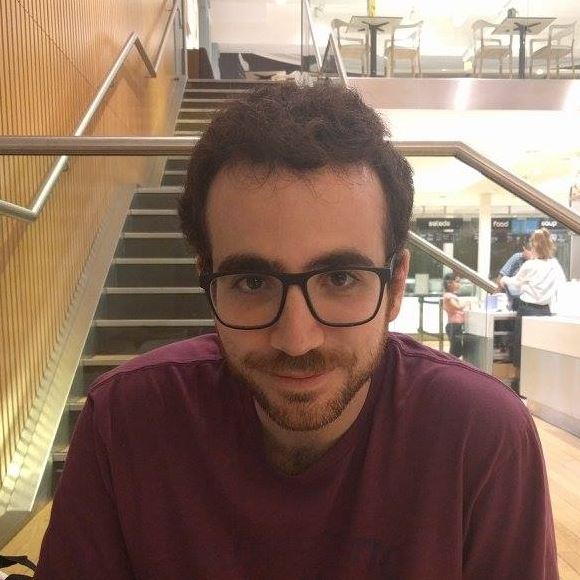}}]{Adria Recasens} is pursuing his PhD in computer vision at the Computer Science and Artificial Intelligence Laboratory (CSAIL) of the Massachusetts Institute of Technology advised by Professor Antonio Torralba. His research interests range on various topics in computer vision and machine learning. He is focusing most of his research on automatic gaze-following. He received a Telecommunications Engineer's Degree and a Mathematics Licentiate Degree from the Universitat Politècnica de Catalunya.
\end{IEEEbiography}

\begin{IEEEbiography}[{\includegraphics[width=1in,height=1.25in,clip,keepaspectratio]{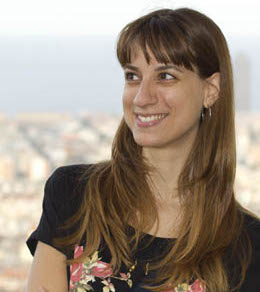}}] {Agata Lapedriza} 
	is an Associate Professor at the Universitat Oberta de Catalunya. She received her MS deegree in Mathematics at the Universitat de Barcelona and her Ph.D. degree in Computer Science at the Computer Vision Center, at the Universitat Autonoma Barcelona. She was working as a visiting researcher in the Computer Science and Artificial Intelligence Lab, at the Massachusetts Institute of Technology (MIT), from 2012 until 2015. Currently she is a visiting researcher at the MIT Medialab, Affective Computing group from September 2017. Her research interests are related to image understanding, scene recognition and characterization, and affective computing.
\end{IEEEbiography}
\vfill
\end{document}